\documentclass[final]{cvpr}
\newcommand{\final}{1}

\usepackage{times}

\ifdefined\siggraph
\usepackage{times}
\fi

\usepackage{color}
\usepackage{ifthen}
\usepackage{float}
\usepackage{alltt}
\usepackage{newlfont} 
\usepackage{wrapfig}
\usepackage{epsfig}
\usepackage{subcaption}
\usepackage{graphicx}
\usepackage{booktabs}
\usepackage{multirow}
\usepackage{makecell}
\usepackage{array}
\usepackage{appendix}
\usepackage{comment}
\usepackage{amsmath}
\usepackage{amssymb}
\usepackage{amsthm}
\usepackage{bm}
\usepackage{textcomp}
\usepackage{mathrsfs} 
\usepackage{url}
\usepackage{gensymb}
\usepackage[symbol]{footmisc}

\newcommand{\nothing}[1]{}

\definecolor{DeltaColor}{rgb}{0.039,0.73,0.71}
\definecolor{SetaColor}{rgb}{0.867, 0.0235, 0.376}
\definecolor{SigmaColor}{rgb}{0.98,0.45,0.0}
\definecolor{RedColor}{rgb}{0.8,0,0}
\definecolor{AlphaColor}{rgb}{0,0,0.8}
\definecolor{BetaColor}{rgb}{0.8,0,0.8}
\definecolor{GammaColor}{rgb}{0.5,0,0.7}
\definecolor{EpsilonColor}{rgb}{0.353,0.725,0.906}
\definecolor{TauColor}{rgb}{0.423,0.235,0.192}

\definecolor{figred}{rgb}{1,0,0}
\definecolor{figgreen}{rgb}{0,0.6,0}
\definecolor{figblue}{rgb}{0,0,1}
\definecolor{figpink}{rgb}{1,0.63,0.63}

\newcommand{\wenbin}[1]{{\color{RedColor} Wenbin: #1 $\qed$}}
\newcommand{\jiabao}[1]{{\color{AlphaColor} Jiabao: #1 $\qed$}}
\newcommand{\yuxin}[1]{{\color{SigmaColor} Yuxin: #1 $\qed$}}
\newcommand{\kui}[1]{{\color{GammaColor} Kui: #1 $\qed$}}

\newcommand{\warning}[1]{{\it\color{red} #1}}
\newcommand{\note}[1]{{\it\color{blue} #1}}

\definecolor{AudioColor}{rgb}{0.56,0.34,0.62}

\definecolor{DeadlineColor}{rgb}{0.9,0.4,0} 
\newcommand{\deadline}[1]{{\bf\color{DeadlineColor} ETA: #1}}

\newcolumntype{C}[1]{>{\centering}m{#1}}

\ifthenelse{\equal{\final}{1}}
{
\renewcommand{\wenbin}[1]{}
\renewcommand{\jiabao}[1]{}
\renewcommand{\yuxin}[1]{}
\renewcommand{\kui}[1]{}
\renewcommand{\warning}[1]{}
\renewcommand{\note}[1]{}
\renewcommand{\deadline}[1]{}

}
{}

\newcounter{pccount}
\floatstyle{plain}

\newcommand{\filename}[1]{\url{#1}}
\newcommand{\foldername}[1]{\url{#1}}

\newtheorem{corollary}{Corollary}[section]

\usepackage[pagebackref=true,breaklinks=true,colorlinks,bookmarks=false]{hyperref}

\def\cvprPaperID{6656} 
\def\confYear{CVPR 2021}

\graphicspath{
}

\begin{document}
%
\title{Sign-Agnostic Implicit Learning of Surface Self-Similarities for Shape Modeling and Reconstruction from Raw Point Clouds}

\author{
        Wenbin Zhao$^{1}$\thanks{Equal contribution}, 
        Jiabao Lei$^{1*}$,
        Yuxin Wen$^{1}$, 
        Jianguo Zhang$^{4}$,
        Kui Jia$^{123}$\thanks{Correspondence to Kui Jia $<$kuijia@scut.edu.cn$>$} \\
	$^1$South China University of Technology. $^2$Pazhou Lab. $^3$Peng Cheng Lab. \\
	$^4$Department of Computer Science and Engineering, Southern University of Science and Technology. \\
	{\tt\small  \{eemszhaowb,eejblei,wen.yuxin\}@mail.scut.edu.cn},\\ {\tt\small zhangjg@sustech.edu.cn}, {\tt\small kuijia@scut.edu.cn}
}

\maketitle

\pagestyle{empty}  
\thispagestyle{empty} 

\begin{abstract}
Shape modeling and reconstruction from raw point clouds of objects stand as a fundamental challenge in vision and graphics research.
Classical methods consider analytic shape priors; however, their performance is degraded when the scanned points deviate from the ideal conditions of cleanness and completeness.
Important progress has been recently made by data-driven approaches, which learn global and/or local models of implicit surface representations from auxiliary sets of training shapes.
Motivated from a universal phenomenon that self-similar shape patterns of local surface patches repeat across the entire surface of an object, we aim to push forward the data-driven strategies and propose to learn a local implicit surface network for a shared, adaptive modeling of the entire surface for a direct surface reconstruction from raw point cloud; we also enhance the leveraging of surface self-similarities by improving correlations among the optimized latent codes of individual surface patches.
Given that orientations of raw points could be unavailable or noisy, we extend sign-agnostic learning into our local implicit model, which enables our recovery of signed implicit fields of local surfaces from the unsigned inputs.
We term our framework as Sign-Agnostic Implicit Learning of Surface Self-Similarities (SAIL-S3).
With a global post-optimization of local sign flipping, SAIL-S3 is able to directly model raw, un-oriented point clouds and reconstruct high-quality object surfaces.
Experiments show its superiority over existing methods.
\end{abstract}

\vspace{-0.5cm}
\section{Introduction}
\label{sec:intro}
\vspace{-0.1cm}

\kui{
1. It is better to present an Algorithm of SAIL-S3

2. We may choose to present a few failure cases, by comparing with other methods, possibly in supplementary material.
}

\begin{figure}[h]
    \begin{center}
        \includegraphics[width=0.45\textwidth]{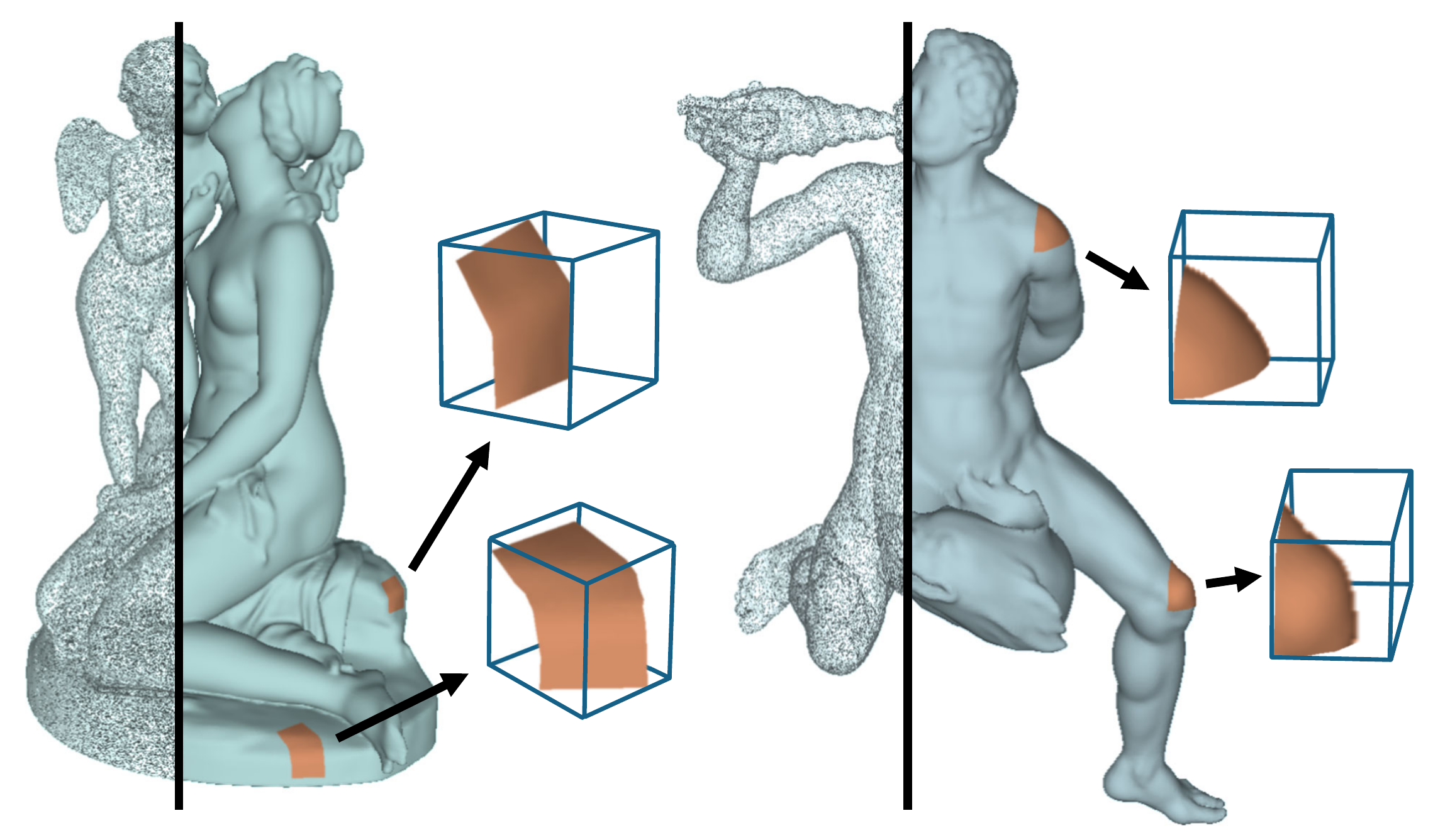}
    \end{center}
    \vspace{-0.5cm}
    \caption{
    3D reconstructions from our proposed Sign-Agnostic Implicit Learning of Surface Self-Similarities.
    For each sculpture, we visualize the raw, un-oriented point cloud on the left, and the reconstructed surface on the right.
    The surface is reconstructed by interpolation from the learned local implicit subfields, where we isolate some of their zero-level sets for better understanding.
    }
    \label{fig:teaser}
    \vspace{-0.65cm}
\end{figure}

Surface reconstruction from point clouds is of significance during the course of digitally representing the world around us, especially when we have witnessed the development of scanning devices that makes it easier to acquire point cloud data.
This problem is severely ill-posed \cite{BergerSurvey17}, since there could be infinite solutions of the continuous surface given the discrete approximation of point clouds, especially when the points are noisy, irregularly distributed, and/or incomplete.
As such, proper priors of geometric regularity are necessary to tackle this problem.
Classical methods adopt analytic priors such as local linearity and smoothness \cite{kazhdan2013screened, carr2001reconstruction, alexa2003computing}.
However, performance of these methods is degraded when encountering sensing imperfection, or unavailable of surface normals for the observed points.

More recently, deep neural networks are introduced to learn geometric priors from auxiliary shapes \cite{Groueix2018, Park2019} in a data-driven manner, which have shown their superiority over classical methods.
The pipeline starts from approaches \cite{Park2019, chen2019learning, Mescheder2019} that globally embed a shape into a latent shape space based on auto-encoders. 
Encoding global shape priors might simplify the problem, which, however, is limited in generalizing the learned priors to unseen shapes.
To improve generalization, there has been some attempts \cite{Chen2019, Genova2019, jiang2020local, Chabra2020, Peng2020} that learn local shape priors and model a global shape as configuration of local shape parts. 
Indeed, for surface shapes of a certain object category, decomposing the global modeling into local ones prevents learning priors that are mostly concerned with the category-level shape configuration. 
These methods rely on learning priors from auxiliary training sets. 
This always risks their generalization in cases that testing samples are out of the distributions of the training ones. 
In this work, we aim to close the generalization gap by learning the shape priors directly from the input data themselves.

Our idea is motivated from an arguably universal phenomenon that self-similar shape patterns of local surface patches repeat across an entire object surface. 
Figure \ref{fig:teaser} gives an illustration. 
Such a phenomenon is similar to the self-similarities of local patches in a 2D image, which has motivated a plethora of methods in the literature of image modeling and restoration \cite{NLM, EladTIP06, BM3D}. 


To implement this phenomenon for modeling and reconstructing a surface from raw observed points, 
a challenge remains due to the possibly unreliable surface normals associated with the observed individual points. One may compute approximate ones, which might not be precise enough to support a fine surface recovery especially when points are noisy or scanner information is absent. Learning to predict the surface normals \cite{erler2020points2surf,guerrero2018pcpnet} is not applicable as well, since we may only have the observed points at hand.
To this end, we propose in this paper a novel method, termed \emph{Sign-Agnostic Implicit Learning of Surface Self-Similarities (SAIL-S3)}, for modeling and reconstruction of a continuous surface directly from a raw, un-oriented point cloud. 
We note that the property of surface self-similarities is also used in \cite{hanocka2020point2mesh} to deform an initial mesh, where they implicitly leverage the property by training the mesh deformation network. 
In contrast, our proposed SAIL-S3 is a completely different local framework for sign-agnostic implicit surface modeling.

Specifically, SAIL-S3 is by design a local model that partitions a global implicit surface field into an \emph{adaptive} set of overlapped, local subfields, each of which is expected to cover a surface patch. 
We leverage the property of surface self-similarities by incorporating the following designs into SAIL-S3: (1) we use a shared implicit model to learn these subfields, while allowing the individual latent representations of local subfields to be freely optimized, and (2) we use a learning objective that promotes correlated latent representations when their corresponding surface patches are of similar shape (cf. Section \ref{sec:method_self_similarities}). 
We extend sign-agnostic learning \cite{Atzmon2020_SAL} into our local framework, and propose provably model initialization that outputs a \emph{signed} solution of implicit field function given the \emph{unsigned} learning objective (cf. Section \ref{sec:method_sign_agnostic}). 
The signed solutions of local implicit subfields may not be consistent globally. 
With a global post-optimization of local sign flipping, SAIL-S3 is able to directly model raw, un-oriented point clouds and reconstructs high-quality object surfaces (cf. Section \ref{sec:method_post_opt}). 
We conduct thorough experiments on the objects from ShapeNet \cite{chang2015shapenet} and Threedscans \cite{threedscans} datasets. 
They include object instances with natural and complex topologies. 
Experiments show that given no auxiliary training set, our proposed SAIL-S3 outperforms existing methods in terms of reconstructing smooth and sharp surfaces, even though the comparative learning based methods use auxiliary training shapes. 
Robustness tests with noisy inputs again confirm the efficacy of our proposed method.
We finally summarize our technical contributions as follows.
\begin{itemize}
\vspace{-0.1cm}
\item We propose a novel method of \emph{SAIL-S3} for surface modeling and reconstruction from raw, un-oriented point clouds. The method learns self-adaptive shape priors by implementing a universal phenomenon that an object surface contains self-similar shape patterns of local surface patches.
\vspace{-0.15cm}
\item SAIL-S3 uses adaptively learned local implicit functions to model the global implicit surface field. We extend sign-agnostic learning into the local SAIL-S3 framework, by proposing provably model initializations that can be optimized to produce \emph{signed} solutions of local implicit function from the \emph{unsigned} learning objective.
\vspace{-0.15cm}
\item With a global post-optimization of local sign flipping, SAIL-S3 is able to directly model raw, un-oriented point clouds and reconstructs high-quality surfaces of objects. Experiments demonstrate its superiority over existing methods.
\end{itemize}

\section{Related works}
In this section, we briefly review existing methods for surface modeling and reconstruction from raw point clouds.
We focus on those works closely related to the present one.

\vspace{0.1cm}
\noindent\textbf{Classical Methods Using Analytic Shape Priors -- }
There have been a number of analytic priors proposed in the literature.
Representative ones include Screened Poisson Surface Reconstruction (SPSR) \cite{kazhdan2013screened}, Radius Basis Functions (RBF) \cite{carr2001reconstruction}, and Moving Least Squares (MLS) \cite{levin2004mesh}.
SPSR a kind of method based on global surface smoothness priors.
It casts the reconstruction as a spatial Poisson problem and solves it in the frequency domain.
However, it relies on oriented normals of surface points.
Likewise, RBF is also based on global surface smoothness priors.
It produces reconstruction through a linear combination of radially symmetric basis functions.
MLS \cite{levin2004mesh} directly approximates the input points as spatially-varying low-degree polynomials, which adopts local surface smoothness as its priors.

\vspace{0.1cm}
\noindent\textbf{Neural Priors for Explicit Surface Modeling  -- }
Given observed points, deep neural networks are recently proposed to encode the points as a latent representation, and then decode it explicitly as a surface mesh.
Among these methods, AtlasNet \cite{Groueix2018} defines the surface as a set of atlas charts, and trains a network to deform their vertices to form a complete surface mesh.
Subsequent methods \cite{wang2018pixel2mesh,hanocka2019meshcnn, tang2019skeleton, tang2020skeletonnet} extend AtlasNet by deforming a single, initial mesh.
Note that the property of self-similarities is also used in \cite{hanocka2020point2mesh} to deform an initial mesh, where they assume these self-similarities are implicitly leveraged by training the network for surface deformation.
However, mesh deformation cannot change surface topologies, and it is difficult for such method to generate surface of complex topologies.
As a remedy, topology modification is proposed in \cite{Pan2019} by pruning edges and faces during the deformation process.
In general, such methods of explicit mesh deformation perform worse than those reconstructing a surface by learning deep implicit fields.

\vspace{0.1cm}
\noindent\textbf{Neural Priors for Implicit Surface Modeling  -- }
More recently, learning deep networks as implicit surface fields is found to be an effective approach for modeling continuous surface \cite{Park2019, chen2019learning, Mescheder2019}, which should be extracted via \cite{lorensen1987marching, lei2020am}.
They typically learn a global surface field of Signed Distance Function (SDF) \cite{Park2019} or occupancy \cite{Mescheder2019}.
Subsequent methods \cite{Chen2019, Genova2019, jiang2020local, Chabra2020, Peng2020} extend them as local implicit models for modeling local surface patches.
For example, BAE-NET \cite{Chen2019} adopts branched decoders for adaptively modeling surface parts; Deep Local Shape \cite{Chabra2020} and Convolutional OccNet \cite{Peng2020} utilize a 3D grid of voxels to represent an SDF or occupancy field, which is memory-expensive; the method \cite{jiang2020local} avoids voxel-based SDF representations, however, it still requires auxiliary shapes for model training.

\vspace{0.1cm}
\noindent\textbf{Sign-Agnostic Surface Modeling  -- }
Practically scanned raw points are usually short of oriented normals.
Analytic computation can only give approximate results.
It is thus appealing to model the raw points in a sign-agnostic manner \cite{Atzmon2020_SAL,Atzmon2020_SALD}.
Atzmon and Lipman study this problem of Sign-Agnostic Learning (SAL).
They propose unsigned objectives, which, given proper initialization of network weights, can produce signed solutions of implicit functions.
Original SAL works with global shape modeling.
In this work, we extend SAL into our local framework, and propose the corresponding network initialization.

\vspace{-0.2cm}
\section{Problem Statement}
\label{SecProblem}
\vspace{-0.1cm}

Given a set of observed points $\mathcal{P} = \{ \bm{p}_i \in \mathbb{R}^3 \}_{i=1}^n$ that represents a discrete sampling of an underlying object surface $\mathcal{S}$, we study a fundamental problem of modeling $\mathcal{S}$ and reconstruct it from the observed $\mathcal{P}$ \cite{BergerSurvey17}.
The problem is severely ill-posed, since there could be infinitely many solutions of the continuous $\mathcal{S}$ given the discrete approximation $\mathcal{P}$; it becomes even more difficult considering that $\mathcal{P}$ may be obtained from practical scanning, and due to imperfection of sensing, the scanned points may be noisy, irregularly distributed, and/or incomplete.
As such, proper priors of geometric regularity is to be imposed in order to recover meaningful solutions.
Classical methods adopt analytic priors such as local linearity and smoothness \cite{kazhdan2013screened, carr2001reconstruction, alexa2003computing}; however, their performance degrades with the increased levels of sensing imperfection.
These methods usually require the availability of surface normals for the observed points, which, however, are either unavailable or cannot be computed accurately.
It is recently shown that learned neural priors from an auxiliary set of training shapes provide strong regularization on the recovery of continuous surface shapes, particularly those based implicit models, \eg, SDF \cite{Park2019} or occupancy field \cite{Mescheder2019, chen2019learning}, and those extending these models as local ones \cite{erler2020points2surf, jiang2020local}.
In this work, we also consider modeling the observed $\mathcal{P}$ with implicit SDFs.
We improve over existing methods by learning directly from $\mathcal{P}$ itself, without relying on the auxiliary training set.

\vspace{-0.2cm}
\section{The Proposed Method}
\label{SecLocalS3Modeling}
\vspace{-0.1cm}
\begin{figure}
    \centering
    \includegraphics[width=0.46\textwidth]{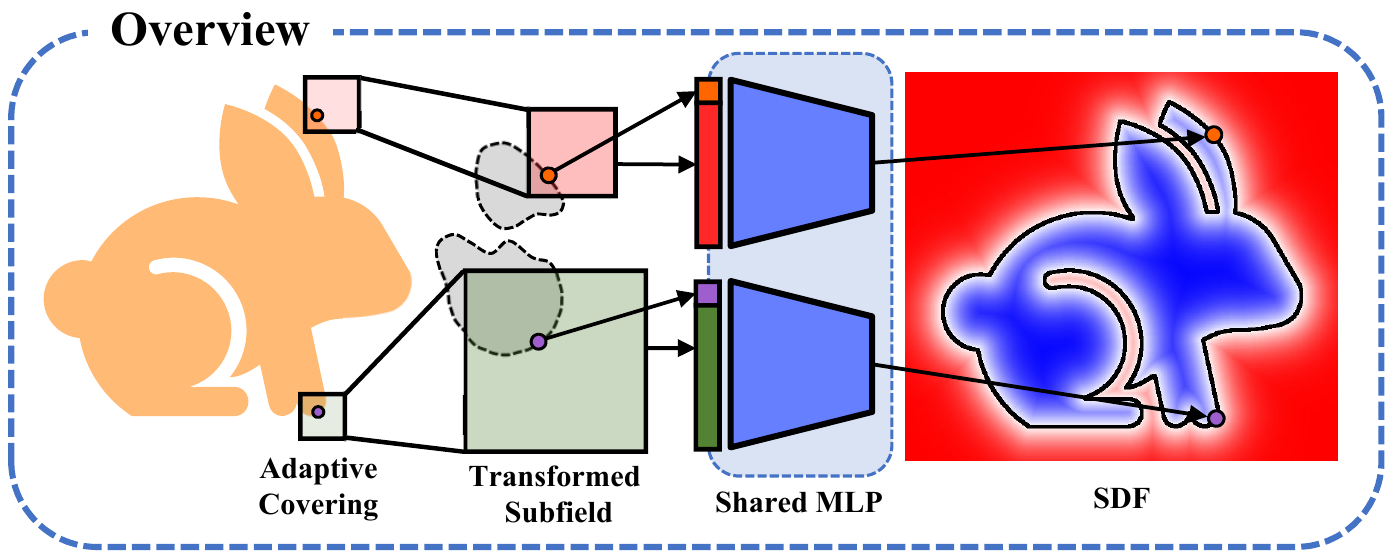}
    \caption{
    An illustration of our proposed Sign-Agnostic Implicit Learning of Surface Self-Similarities (SAIL-S3). 
    }
    \label{fig:pipeline}
    \vspace{-0.5cm}
\end{figure}

Our proposed method is primarily motivated from a universal phenomenon that self-similar shape patterns of local surface patches repeat across an entire object surface.
Figure \ref{fig:teaser} gives an illustration.
Such a phenomenon is similar to the self-similarities of local patches in a 2D image, which has motivated a plethora of methods in the literature of image modeling and restoration, including the representative non-local means \cite{NLM}, self-adaptive dictionary learning \cite{EladTIP06}, and BM3D \cite{BM3D}; they have successfully shown that clean images can be restored from distorted images themselves.
Motivated from the conceptually similar surface self-similarities, we aim to address the problem stated in Section \ref{SecProblem}, and propose a novel method termed \emph{Sign-Agnostic Implicit Learning of Surface Self-Similarities (SAIL-S3)}.
An illustration of it is shown in Figure \ref{fig:pipeline}.
Details are presented as follows.

\subsection{Local Implicit Modeling of Surface Self-Similarities}
\label{sec:method_self_similarities}

Denote $\mathcal{G} \subset \mathbb{R}^3$ as an implicit field of SDF, whose zero-level set represents the underlying surface $\mathcal{S}$ from which the observed points in $\mathcal{P}$ are sampled.
Let $\hat{f}: \mathbb{R}^3\times \mathbb{R}^d \rightarrow \mathbb{R}$ be the implicit model of $\mathcal{G}$; it takes as input a sampled point $\bm{q} \in \mathbb{R}^3$ in the 3D space and a latent representation $\hat{\bm{z}} \in \mathbb{R}^d$ encoding the surface, and outputs a value of signed distance between $\bm{q}$ and $\mathcal{S}$.
Instead of obtaining $\hat{\bm{z}}$ from any learned encoder, we follow \cite{Park2019} and resort to latent code optimization to fit with the observed $\mathcal{P}$.


The pre-assumed property of surface self-similarities suggests that some local surface patches of $\mathcal{S}$ are similar in terms of shape pattern.
Our design for leveraging this property has the following two ingredients.
\begin{itemize}
\vspace{-0.25cm}
\item  Instead of directly learning $\hat{f}$ for the global field $\mathcal{G}$, we consider a number $N$ of overlapped, local implicit subfields $\{ \mathcal{F}_i \}_{i=1}^N$, each of which is responsible for a surface patch $\mathcal{S}_{\mathcal{F}_i}$; we use a shared implicit model $f_{\bm{\theta}}: \mathbb{R}^3\times \mathbb{R}^d \rightarrow \mathbb{R}$, parameterized by $\bm{\theta}$, to learn these subfields, while allowing the individual latent representations $\{ \bm{z}_i \in \mathbb{R}^d \}_{i=1}^N$ to be freely optimized.
\vspace{-0.3cm}
\item  We use a learning objective that promotes correlated latent representations when their corresponding surface patches are of similar shape; for example, a pair of correlated $\bm{z}_i$ and $\bm{z}_j$ outputs, through $f_{\bm{\theta}}$, implicit subfields $\mathcal{F}_i$ and $\mathcal{F}_j$ whose zero-level sets represent similar shapes.
\end{itemize}
\vspace{-0.2cm}
While the latter ingredient is an explicit design to promote surface self-similarities, the former one implicitly does so by decoding the implicit subfield with the shared model $f_{\bm{\theta}}$.
In this work, we implement $f_{\bm{\theta}}$ as a network of multi-layer perceptron (MLP). Each subfield $\mathcal{F}_i$, $i \in \{1, \dots, N\}$, is centered at $\bm{c}_i \in \mathbb{R}^3$ and covers a local volume of size $a_i \times a_i \times a_i$. Both $\mathcal{C} = \{\bm{c}_i\}_{i=1}^N$ and $\mathcal{A} = \{a_i\}_{i=1}^N$ are learnable parameters, which determine how $\{ \mathcal{F}_i \}_{i=1}^N$ distribute in the global $\mathcal{G}$; we expect their optimizations to make each of $\{ \mathcal{F}_i \}_{i=1}^N$ cover a patch of the surface $\mathcal{S}$ (cf. Section \ref{SecAdaptFieldCovering} for the details).
Let $\mathcal{Z} = \{ \bm{z}_i \in \mathbb{R}^d \}_{i=1}^N$.\jiabao{this sentence seems a bit obtrusive in the context?}
SAIL-S3 is formally to learn the implicit SDF $f_{\bm{\theta}}: \mathbb{R}^3\times \mathbb{R}^d \rightarrow \mathbb{R}$, parameterized by $(\bm{\theta}, \mathcal{Z}, \mathcal{C}, \mathcal{A})$ , by fitting to the observed $\mathcal{P}$.
We note that defining implicits over cubes practically supports more convenient space partitioning (cf. Section \ref{SecAdaptFieldCovering}) and smoothing of the results in overlapped regions (cf. Section \ref{sec:method_infere}).
We present the sign-agnostic implicit learning of SAIL-S3 as follows.

\vspace{-0.1cm}
\subsection{Sign-Agnostic Local Implicit Learning}
\label{sec:method_sign_agnostic}
\vspace{-0.1cm}
An important challenge for shape modeling of raw point clouds is the possibly unreliable surface normals associated with the observed points. Approximate ones may be estimated from $\mathcal{P}$ via either covariance analysis \cite{hoppe1992surface} or learning-based methods \cite{erler2020points2surf,guerrero2018pcpnet}; however, their performances usually degrade especially when inputs are noisy or camera information is absent, which is hard to support accurate surface recovery.
SAL \cite{Atzmon2020_SAL} is a promising solution to cope with the issue; it is proposed to learn a global implicit model to reconstruct an entire object surface, with no requirement on the availability of surface normals.
However, it remains absent for how to use it to model a surface as a collection of local implicit subfields.
In this work, we extend the SAL technique \cite{Atzmon2020_SAL} into our local framework of SAIL-S3, as follows.

We first present the extension in a local implicit subfield $\mathcal{F}$.
Before that, for any point $\bm{q}$ sampled in the global $\mathcal{G}$, we compute its \emph{unsigned} distance $s(\bm{q}) \in \mathbb{R}^{+}$ to the surface $\mathcal{S}$ of interest approximately as
\begin{equation}\label{EqnUnsignedDist}
\setlength\abovedisplayskip{3pt}
\setlength\belowdisplayskip{3pt}
s(\bm{q}) = \| \bm{q} - \bm{p} \|_2 \ \ \textrm{s.t.} \ \ \bm{p} = \arg\min_{\bm{p}' \in \mathcal{P}} \| \bm{q} - \bm{p}' \|_2 .
\end{equation}
The distance (\ref{EqnUnsignedDist}) approaches the true one when the number $n$ of points in $\mathcal{P}$ goes to infinity.
In practice, it would provide us an approximate supervision signal for learning the implicit model.
Its unsigned nature relaxes the requirement on the knowledge of local surface orientations.

Assume that $\mathcal{F}$ contains a set $\mathcal{P}_{\mathcal{F}}$ of observed points in a local neighborhood of $\mathcal{P}$.
We normalize any $\bm{p} \in \mathcal{P}_{\mathcal{F}} \subset \mathcal{G}$ as $\bar{\bm{p}} = (\bm{p} -\bm{c})/a$ in a local coordinate system of the subfield $\mathcal{F}$, where $\bm{c}$ and $a$ are the learnable center and side length.
We applies the same to any sampled point $\bm{q} \in \mathcal{F} \subset \mathcal{G}$, resulting in $\bar{\bm{q}}$ after coordinate offset and scaling.
We thus have the unsigned distance $s(\bar{\bm{q}}) = s(\bm{q}) / a$ inside $\mathcal{F}$.  Given the supervision from $\{ \bar{\bm{q}} \in \mathcal{F} \}$, SAL \cite{Atzmon2020_SAL} aims to find a \emph{signed} solution of the implicit function $f_{\bm{\theta}}$, by solving the following \emph{unsigned}, bi-level optimization problem
\begin{eqnarray}\label{EqnLocalUnsignedObj}
\min_{ \bm{\theta}, \bm{z}}  \sum_{\bar{\bm{q}} \in \mathcal{F}} \left| \left| f_{\bm{\theta}}(\bar{\bm{q}}, \bm{z}) \right| - s(\bar{\bm{q}}) \right| \ \textrm{s.t.} \ \ s(\bar{\bm{q}}) = s(\bm{q}) / a ,
\end{eqnarray}
where we have temporarily assumed that $a$ and $\bm{c}$ for the subfield $\mathcal{F}$ are fixed, and $\{ \bar{\bm{q}} \in \mathcal{F} \}$ (equivalently, $\{ \bm{q} \in \mathcal{F} \}$) are usually sampled around individual $\bm{p} \in \mathcal{P}_{\mathcal{F}}$ with densities inversely proportional to the distances.
A signed solution of (\ref{EqnLocalUnsignedObj}) means that for an optimal $\bm{z}^{*}$, we have $f_{\bm{\theta}^{*}}(\bar{\bm{q}}, \bm{z}^{*})  > 0$ when $\bar{\bm{q}}$ lies by one side of the local surface $\mathcal{S}_{\mathcal{F}}$, and $f_{\bm{\theta}^{*}}(\bar{\bm{q}}, \bm{z}^{*}) < 0$ otherwise; in other words, it produces signed distances to the surface even though the supervision is unsigned.
However, this is not always guaranteed given that a flipped sign of $f_{\bm{\theta}}(\bar{\bm{q}}, \bm{z})$ does not change its absolute value, and consequently the loss (\ref{EqnLocalUnsignedObj}).
The problem becomes even more involved when coupled with the simultaneous optimization of $\bm{z}$.

Fortunately, it is suggested in \cite{Atzmon2020_SAL} that a proper setting of network weights $\bm{\theta}^0$ would initialize a latent code-free function $f_{\bm{\theta}^0}(\cdot)$ as a signed function, and it is also empirically observed that optimization from such an initialization is stably in the signed local minima, without going across loss barriers to the unsigned solutions.
In this work, we extend the weight initialization scheme in \cite{Atzmon2020_SAL} for learning a signed solution of local implicit function from the loss (\ref{EqnLocalUnsignedObj}), as presented shortly.

\begin{figure}[h]
    \begin{center}
        \includegraphics[width=0.46\textwidth]{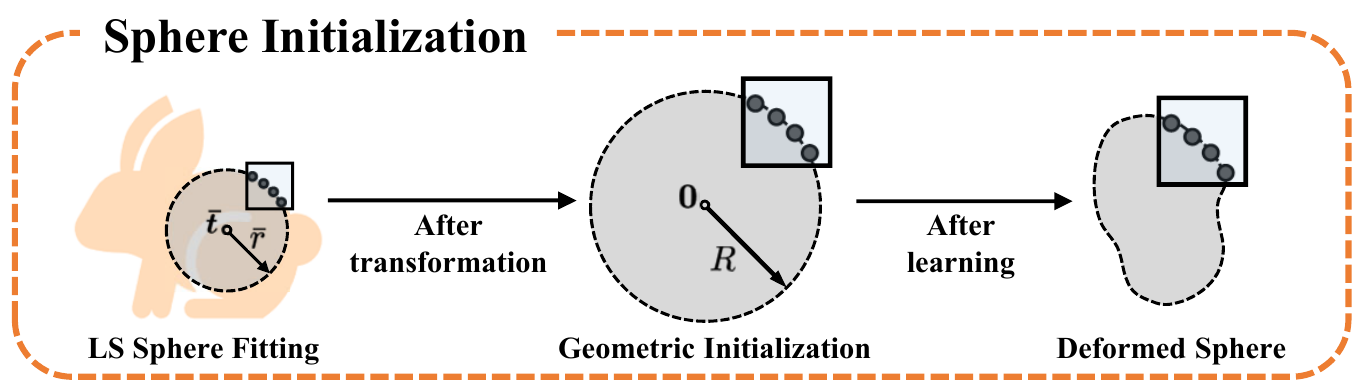}
    \end{center}
    \vspace{-0.5cm}
    \caption{An illustration on our model initialization.
    }
    \label{fig:sphere}
    \vspace{-0.4cm}
\end{figure}

\vspace{0.1cm}
\noindent\textbf{A Proper Model Initialization for Signed Solutions} For the observed points $\{ \bar{\bm{p}} \in \mathcal{P}_{\mathcal{F}} \}$ in $\mathcal{F}$, we first use least squares to fit them with a surface patch on a radius-$\bar{r}$ sphere centered at $\bar{\bm{t}}$ (cf. the supplementary material for more details). The following corollary shows a scheme of initializing $\bm{\theta}^0$ and $\bm{z}^0$ that guarantees $f_{\bm{\theta}^0}(\bar{\bm{q}}, \bm{z}^0) \approx \| \bar{\bm{q}} - \bar{\bm{t}} \| - \bar{r}$; in other words, $f_{\bm{\theta}^0}(\cdot, \bm{z}^0)$ is initialized as a signed distance function whose zero-level set is approximately a $\bar{r}$-radius sphere centered at $\bar{\bm{t}}$, and the sphere approximately covers the local point set $\mathcal{P}_{\mathcal{F}}$.
We experimentally find that our results are relatively stable w.r.t. LS sphere initialization.
Figure \ref{fig:sphere} gives an illustration.

\vspace{-0.1cm}
\begin{corollary}
\label{InitCol}
Let $f:\mathbb{R}^{3+d}\to\mathbb{R}$ be an $l$-layer MLP with ReLU activation $\nu$.
That is, $f(\bm{p},\bm{z})=\bm{w}^T\nu(\bm{W}^{l}(\cdots\nu(\bm{W}_{\bm{p}}^{1}\bm{p}+\bm{W}_{\bm{z}}^{1}\bm{z}+\bm{b}^{1}))+\bm{b}^{l})+c$, where $\bm{W}_{\bm{p}}^{1}\in\mathbb{R}^{d^{1}_{\textrm{out}}\times 3}$ and $\bm{W}_{\bm{z}}^{1}\in\mathbb{R}^{d^{1}_{\textrm{out}}\times d}$ denote the weight matrices of the first layer, and $\bm{b}^{1}\in\mathbb{R}^{d^{1}_{\textrm{out}}}$ denotes the bias; $\bm{W}^{i}\in\mathbb{R}^{d^{i}_{\textrm{out}}\times d^{i-1}_{\textrm{out}}}$ and $\bm{b}^{i}\in\mathbb{R}^{d^{i}_{\textrm{out}}}$ denote parameters of the $i^{th}$ layer; $\bm{w}\in\mathbb{R}^{d_{\textrm{out}}^{l}}$ and $c\in\mathbb{R}$ are parameters of the last layer; $\bm{p}\in\mathbb{R}^3$ is the input point, and $\bm{z}\in\mathbb{R}^{d}$ is the latent code, whose elements follow the i.i.d. normal $\mathcal{N}(0,\sigma_{z}^2)$.
Let $\bm{w}=\sqrt{\frac{\pi}{d_{\textrm{out}}^{l}}}\bm{1}$, $c=-\bar{r}$, $\bar{r}>0$, let all entries of $\bm{W}^{i}$ ($2\leq i\leq l$) follow i.i.d. normal $\mathcal{N}(0,\frac{2}{d_{\textrm{out}}^{i}})$, let entries of $\bm{W}_{\bm{p}}^{1}$ follow i.i.d. normal $\mathcal{N}(0,\frac{2}{d_{\textrm{out}}^{1}})$, and let $\bm{b}^i=\bm{0}$ ($2\leq i\leq l$).
If $\bm{W}_{\bm{z}}^{1}=\bm{W}_{\bm{p}}^{1}[\bm{I}\in\mathbb{R}^{3\times 3}, \bm{0}\in\mathbb{R}^{3\times {(d-3)}}]$ and $\bm{b}^1=-\bm{W}_{\bm{p}}^{1}\bar{\bm{t}}$, then $\lim_{\sigma_z \to 0} f(\bm{p},\bm{z}) = \|\bm{p}-\bar{\bm{t}}\| - \bar{r}$.
That is, $f$ is approximately the signed distance function to a 3D sphere of radius $\bar{r}$ centered at $\bar{\bm{t}}$.
\end{corollary}
\noindent The proof to Corollary \ref{InitCol} is provided in the supplementary material.

\vspace{-0.4cm}
\subsubsection{Learning over a Collection of Local Implicit Subfields}
\vspace{-0.15cm}

We have so far presented how to learn an implicit function individually for a local subfield.
For learning over the collection $\{ \mathcal{F}_i \}_{i=1}^N$, we remind that SAIL-S3 shares the function $f_{\bm{\theta}}$ for these subfields, where $\bm{\theta}$ is initialized and optimized for all the $N$ subfields, while the latent codes $\{ \bm{z}_i \}_{i=1}^N$ are adaptively optimized.
This brings an inconsistency when initializing the implicit function separately for each subfield based on Corollary \ref{InitCol}.
In practice, we circumvent this inconsistency by first initializing $\bm{\theta}^0$ as suggested by Corollary \ref{InitCol}, and $\{ \bm{z}_i^0 \}_{i=1}^N$ as the samples drawn from a Gaussian distribution with small standard deviation, which essentially defines the zero-level set of $f_{\bm{\theta}^0}(\cdot, \bm{z}_i^0)$ as a radius-$R$ sphere centered at the origin, where $R$ is a hyper-parameter; for the $i^{th}$ subfield, we then transform its sampled point $\bar{\bm{q}}^i \in \mathcal{F}_i$ as
\begin{equation}\label{EqnTransform4Normalize}
\setlength\abovedisplayskip{4pt}
\setlength\belowdisplayskip{4pt}
\tilde{\bm{q}}^i = R (\bar{\bm{q}}^i - \bar{\bm{t}}_i) / \bar{r}_i  ,
\end{equation}
where we use superscript $i$ in $\bar{\bm{q}}^i$ (and $\tilde{\bm{q}}^i$) to indicate that it is the transformed coordinates of $\bm{q}$ in the $i^{th}$ subfield (note that a sampled $\bm{q}$ may appear in different subfields), and $\bar{\bm{t}}_i$  and $\bar{r}_i$ associated with the subfield $\mathcal{F}_i$ are the center and radius of a sphere obtained by solving a least square fitting problem. 
Note that by (\ref{EqnTransform4Normalize}), any observed point $\bar{\bm{p}} \in \mathcal{P}_{\mathcal{F}_i}$ is transformed in the same way as $\tilde{\bm{p}}$; this means geometrically that each subfield is transformed such that its contained point observations approximately fit with a surface patch on the initialized zero-level sphere of radius $R$.
Figure \ref{fig:sphere} gives the illustration.

Given $\{ \tilde{\bm{q}}^i \in \mathcal{F}_i \}$ sampled from each local subfield, we have the following loss function for modeling the underlying surface $\mathcal{S}$ over the collection $\{ \mathcal{F}_i \}_{i=1}^N $
\begin{equation}\label{EqnLossSAILS3}
\setlength\abovedisplayskip{3pt}
\setlength\belowdisplayskip{3pt}
    \begin{split}
    \mathcal{L}^{\textrm{\tiny Modeling}} (\bm{\theta}, \mathcal{Z}, \mathcal{C}, \mathcal{A}) = \sum_{i=1}^N \sum_{\tilde{\bm{q}}^i \in \mathcal{F}_i} &\left| \left| f_{\bm{\theta}}(\tilde{\bm{q}}^i(\bm{c}_i, a_i), \bm{z}_i) \right| - s(\tilde{\bm{q}}^i) \right|  \\
    &+ \lambda \Vert \bm{Z} \Vert_{*},
    \end{split}
\end{equation}
where $s(\tilde{\bm{q}}^i) = s(\bm{q})R/ (\bar{r}_i a_i)$, and $\bm{Z} = [\bm{z}_1 / \Vert \bm{z}_1 \Vert_2; \cdots; \bm{z}_N / \Vert \bm{z}_N \Vert_2 ]$ is collection of all the normalized latent codes.
We
use nuclear norm penalty to improve correlations among latent codes, which ensures the self-similarities learned from the input itself.
Besides, the above modeling loss can be further improved by leveraging the first derivative of function $f_{\bm{\theta}}(\cdot, \bm{z}_i)$ as detailed in \cite{Atzmon2020_SALD}.
\kui {--- where is the nuclear norm penalty to improve correlations among latent codes? Please revise. Please also check whether the above is correct ---}

\vspace{-0.4cm}
\subsubsection{An Adaptive Field Covering}
\label{SecAdaptFieldCovering}
\vspace{-0.15cm}

Our motivation for modeling $\mathcal{S}$ with local self-similarities expects that each individual subfield $\mathcal{F}$ covers a roughly similar volume of the entire $\mathcal{G}$, and the covering would be evenly distributed along the surface $\mathcal{S}$.
To this end, we initialize parameters $\{ (\bm{c}_i, a_i)\}_{i=1}^N$ of local implicit subfields as follows.
Given the observed $\mathcal{P}$, we use farthest point sampling \cite{eldar1997farthest} of $\mathcal{P}$ to initialize the subfield centers $\{\bm{c}_i\}_{i=1}^N$; we then initialize the covering size of subfield as $a_i = \alpha \min_{j \in \{1, \ldots, N\}/\{i\} } \Vert \bm{c}_i - \bm{c}_j \Vert_2$, where we set $\alpha \geq 1$ such that each observed point in $\mathcal{P}$ is covered by at least one subfield and these subfields have a certain amount of overlapping.
This initialization can roughly meet our expectation; however, directly solving the objective (\ref{EqnLossSAILS3}) may update $\{ (\bm{c}_i, a_i)\}_{i=1}^N$ such that a very few of the subfields are enlarged to cover large portions of the surface, while the remaining ones are moved to cover duplicate surface patches.
To avoid these undesired solutions, we propose the following loss terms to constrain the optimization

\vspace{0.1cm}\noindent\textbf{Volume loss -- }
We prevent undesirable enlarging of individual subfields by penalizing the volume of each subfield
\begin{equation}\label{EqnLossVolume}
\setlength\abovedisplayskip{3pt}
\setlength\belowdisplayskip{3pt}
\mathcal{L}^{\textrm{\tiny Volume}}(\mathcal{A}) =  \sum_{i=1}^N \max\{a_i, 0\} .
\end{equation}

\vspace{0.1cm}\noindent\textbf{Placing loss -- }
Given the constraint from (\ref{EqnLossVolume}), we further encourage an even distribution of the subfields $\{ \mathcal{F}_i \}_{i=1}^N$ on the surface by penalizing the Chamfer Distance \cite{fan2017point} between each observed $\bm{p} \in \mathcal{P}$ and its closest subfield center, i.e.,
\begin{equation}\label{EqnLossCovering}
\setlength\abovedisplayskip{3pt}
\setlength\belowdisplayskip{3pt}
\mathcal{L}^{\textrm{\tiny Placing}}(\mathcal{C}) = \sum_{\bm{p} \in \mathcal{P}} \min_{\bm{c}_i \in \mathcal{C}} \| \bm{p} - \bm{c}_i \|_2^2 + \sum_{\bm{c}_i \in \mathcal{C}} \min_{\bm{p} \in \mathcal{P}} \| \bm{p} - \bm{c}_i \|_2^2.
\end{equation}

\vspace{0.1cm}\noindent\textbf{Covering loss -- }
To cover all the observed points in $\mathcal{P}$, we use the exterior signed distance field to penalize those uncovered point $\bm{p}$, i.e.
\begin{equation}
\setlength\abovedisplayskip{3pt}
\setlength\belowdisplayskip{3pt}
    \mathcal{L}^{\textrm{\tiny Covering}} = \sum_{\substack{\bm{p}\in\mathcal{P} \\ \bm{p}\notin \mathcal{G}}}
    \sqrt{\min_{i\in\{1, \ldots, N \}}{
            \sum_{j=1}^{3} \max\{|\pi_j(\bm{p} - \bm{c}_i)| - a_i, 0\}^2
         }},
\end{equation}
where $\pi_j$ is an operator that selects the $j^{th}$ element from a vector.

\vspace{-0.4cm}
\subsubsection{The Combined Learning objective}
\vspace{-0.15cm}

Given the observed $\mathcal{P}$, we use the following combined objective to reconstruct its underlying surface $\mathcal{S}$ via sign-agnostic implicit learning of surface self-similarities
\begin{equation}\label{EqnLossOverall}
\setlength\abovedisplayskip{3pt}
\setlength\belowdisplayskip{3pt}
\mathcal{L}^{\textrm{\tiny SAIL-S3}}
=
\mathcal{L}^{\textrm{\tiny Modeling}} +
\lambda_1 \mathcal{L}^{\textrm{\tiny Volume}} +
\lambda_2 \mathcal{L}^{\textrm{\tiny Placing}} +
\lambda_3 \mathcal{L}^{\textrm{\tiny Covering}}
\end{equation}
where $\lambda_1$, $\lambda_2$, and $\lambda_3$ are penalty parameters.
Starting from initializations, as presented above, the objective (\ref{EqnLossOverall}) can be optimized simply via stochastic gradient descent.
\jiabao{is it necessary to mention this??}
Penalty parameters $\lambda_1$, $\lambda_2$, and $\lambda_3$ can be found via grid search, and they are practically insensitive to all kinds of experimental settings.


\vspace{-0.10cm}
\subsection{A Global Post-Optimization of Local Sign Flipping}
\label{sec:method_post_opt}
\vspace{-0.1cm}

Solving the objective (\ref{EqnLossOverall}) produces $N$ local implicit functions $f_{\bm{\theta}}(\cdot, \bm{z}_i)$, $i = 1, \dots, N$, which may be used for extraction of their zero-level iso-surfaces.
For any $\bm{q} \in \mathcal{G}$ covered by a pair of neighboring $i^{th}$ and $j^{th}$ functions, however, their function evaluations may not be consistent in their signs because each subfield is optimized individually.
Motivated by the procedure of reorienting point cloud in \cite{hoppe1992surface}, we propose a post-optimization of local sign flipping to address this issue.
More specifically, we treat each local implicit subfield $\mathcal{F}$ as a vertex, and construct a connected, undirected graph over the vertices as $G = (\mathcal{V}, \mathcal{E})$, where $\mathcal{V} = \{v_1, \dots, v_N\}$ and an edge $e_{i,j} \in \mathcal{E}$ is connected once the subfields $\mathcal{F}_i$ and $\mathcal{F}_j$ have an overlapped region in the field $\mathcal{G}$.
We associate each edge $e_{i, j}$ with a pair of weights defined as
\begin{eqnarray}\label{EqnEdgeWeight}
\setlength\abovedisplayskip{3pt}
\setlength\belowdisplayskip{3pt}
w_{i, j}^1(e_{i,j}) = \sum_{\bm{q} \in \mathcal{F}_i\cap \mathcal{F}_j } \left| f_{\bm{\theta}}(\tilde{\bm{q}}^i, \bm{z}_i) - f_{\bm{\theta}}(\tilde{\bm{q}}^j, \bm{z}_j) \right| , \nonumber \\
w_{i, j}^0(e_{i,j}) = \sum_{\bm{q} \in \mathcal{F}_i\cap \mathcal{F}_j } \left| f_{\bm{\theta}}(\tilde{\bm{q}}^i, \bm{z}_i) + f_{\bm{\theta}}(\tilde{\bm{q}}^j, \bm{z}_j) \right| . \nonumber
\end{eqnarray}
This gives two weight sets $\mathcal{W}^1$ and $\mathcal{W}^0$ with $|\mathcal{W}^0| = |\mathcal{W}^1| = |\mathcal{E}|$, and we write $\mathcal{W} = \mathcal{W}^0 \cup \mathcal{W}^1$.
We endow each vertex $v \in \mathcal{V}$ with a sign variable $h(v) \in \{1, -1\}$, and use minimum spanning tree (MST) to determine $\{ h(v_i) \}_{i=1}^N$ (details are given in the supplementary material).

We finally flip local implicit functions as $h(v_i) f_{\bm{\theta}}(\cdot, \bm{z}_i)$ (or equivalently, $h(\mathcal{F}_i) f_{\bm{\theta}}(\cdot, \bm{z}_i)$), $i = 1, \dots, N$.
Our MST is based on the Prim's algorithm \cite{prim1957shortest}, which is guaranteed to find a solution of minimal cost; we empirically observe that it works well in practice.

\vspace{-0.1cm}
\subsection{Inference via Interpolation of Local Fields}
\label{sec:method_infere}
\vspace{-0.05cm}
After post-optimizing the local sign as described in Section \ref{sec:method_post_opt}, the predictions for some point $\bm{q} \in \mathcal{G}$ are not expected to be consistent since each prediction may have slight error.
We extend trilinear interpolation to the case of arbitrary number of overlapping regions.
The weights for interpolating are calculated by the following.
For any $\bm{q} \in \mathcal{G}$ falling in a number $M$ of overlapped subfields $\{ \mathcal{F}_j \}_{j=1}^M$, we evaluate its signed distance to the underlying surface $\mathcal{S}$ as the following averaged one
\begin{eqnarray}\label{EqnWeightedInference}
\setlength\abovedisplayskip{3pt}
\setlength\belowdisplayskip{3pt}
\hat{f}(\bm{q}) = \sum_{j=1}^M \omega_j(\bm{q}) \cdot h(\mathcal{F}_j) \cdot \frac{\bar{r}_ja_j}{R} f_{\bm{\theta}}(\tilde{\bm{q}}^j, \bm{z}_j)  \\
\textrm{s.t.} \omega_j(\bm{q}) =
\frac{\left| \max\limits_{k\in\{1,2,3\}} |\pi_k(\bm{q} - \bm{c}_j)| - a_j \right|}
{\sum_{j'=1}^{M} \left|\max\limits_{k\in\{1,2,3\}} |\pi_k(\bm{q} - \bm{c}_{j'})| - a_{j'}\right| }, \nonumber
\end{eqnarray}

Instead of directly using the inverse distance weighted average method mentioned in \cite{qi2017pointnet++}, the proposed interpolation ensures truly smooth transitions between different overlapped regions without discontinuity of the first kind when switching the neighbors.
Such an inference via interpolation has the benefit of smoothing out the less consistent local subfields inferred from individual functions.
We practically observe that training procedure will slow down if applying (\ref{EqnWeightedInference}) during training, compared to simply adopting (\ref{EqnLossSAILS3}). Experiments confirmed the efficacy of our choice to use (\ref{EqnWeightedInference}) as a post-processing method during inference.

Given the signed distance evaluation (\ref{EqnWeightedInference}), we finally use marching cubes \cite{lorensen1987marching} to extract the zero-level set that reconstructs the surface $\mathcal{S}$.


\nothing{
Given an arbitrary raw point cloud $\mathcal{P}$ without corresponding normals, our goal is to reconstruct the underlying object surface $\mathcal{S}$ of it.
To deal with arbitrary raw point cloud, we adopt the features learned from local self-similarities, because though different shapes across different categories and scenes have vastly different geometric forms and topologies at a global scale, they usually share similar features at a certain local scale, making the learning of local similarities feasible.
For a specific object, especially the object of complex shape or high repeatability, such local features could be sufficiently learnt from the object itself, without introducing other auxiliary data.
The features learned from self-similarities will be further more consistent with the characteristic of the object.
A powerful self-prior learned from self-similarities even excels at modeling natural shapes, where noisy and unnatural shapes are neglected due to the their unstructured peculiarity.
What's more, to deal with raw point cloud without normals locally, we further introduce sign-agnostic implicit learning specifically designed for object partition.
These all together form our proposed method.

\subsection{Method Overview}

The schematic overview of our approach is depicted in Figure \ref{fig:overview}.
Given a raw point cloud without normals, we firstly partition the space of the object into different local cells.
And then we learn geometrical feature from the points in each cell via implicit auto-decoder, whose parameters $\bm{\theta}$ are shared and latent shape-code $\bm{z}$ are independent.
During learning, the position of the cell are adjustable to improve the effective of surface learning.
However, the learning on the surface is sign-agnostic, and thus predictions between two adjacent cells on the internal/external of object might be discontinuous.
We need a post-processing method to adjust the sign for each cell globally to make the internal/external information consistent across cells.
What's more, directly unite all the cells might induce discontinuities across cells' boundaries, and thus we further use another post-processing method to interpolate implicit values predicted by the adjacent local networks to generate smoothing surface.
Finally, we adopt the de-facto standard algorithm of Marching Cubes to extract the manifold iso-surface of the interpolated implicit field.
In the following part of this section, we will describe each operation in detail.
\subsection{Space Partition}\label{sec:cells}
We adopt self-adapting cells to partition the 3D space of given point clouds.
The cells $\mathcal{C}$ is defined as cubes of local region for the implicit auto-decoder to learned from.
Self-adapting here means that the centers $\{\bm{c}_i\}_{i=1}^n$, $\bm{c}_i \in \mathbb{R}^3$, and scales $\{s_i\}_{i=1}^n$, $s_i \in \mathbb{R}$, of the cells are not defined in advance, but adjustable accordingly, \ie, the concept of local regions depends on the given input, which would prevent unreasonable receptive fields that are too complex or excessively lacking semantic information to learn from. 
To be more specific, given a point cloud, we firstly down-sample it into $n$ points via farthest point sampling method following \cite{}, where the number of points is determined by the desired number of cells.
We take the sampled points as the initialized centers of the cells, and the $s^0$ as the initialized scales.
During training process, we will adjust centers $\{\bm{c}_i\}_{i=1}^n$ and scales $\{s_i\}_{i=1}^n$ accordingly.
With such deformations for space partition, we want to define effective grid cells, which is helpful for the local surface learning and post-processing.
More details are presented in Section \ref{sec:cells_adapting}.

\subsection{Local Surface Learning} \label{sec:local_surface_learning}
To deal with local surface, we introduce local implicit auto-decoder, which is a simple fully connected neural network that takes in a latent shape-code as well as a 3D point coordinate, and outputs the corresponding implicit function value for the point.
In this section, we will introduce the detail of how we perform local surface learning during training process.

\subsubsection{Point-wise Sign Agnostic Fitting Loss}
\label{sec:SAF}
Both the latent shape-code $\{\bm{z}_i\}_{i=1}^n$, the network parameters $\bm{\theta}$, the centers $\{\bm{c}_i\}_{i=1}^n$ and the scales $\{s_i\}_{i=1}^n$ of the cells are trained with the distance field so that the network learns a continuous decision boundary of the encoded surface.
Formally, the \emph{Sign-Agnostic Fitting} (SAF) loss is given as
\begin{equation}
    \mathcal{L}_{\text{SAF}}(\mathcal{P}) = \sum_{\substack{i\in\{1, \ldots, |\mathcal{C}|\}, \\ j\in\{1, \ldots, |\hat{\mathcal{P}}_{\mathcal{C}_i}|\}}} \bigg|\, \big|f(\hat{\bm{p}}_{i,j}, \bm{z}_i;\bm{\theta})\big| - s(\hat{\bm{p}}_{i,j}) \,\bigg|,
\label{eq:loss_SAF}
\end{equation}
where $\mathcal{C}_i$ denotes the $i^{th}$ cell among the set of cells $\mathcal{C}$, $\hat{\mathcal{P}}_{\mathcal{C}_i}$ denotes the set of points uniformly sampled in such cell, and $\hat{\bm{p}}_{i,j}$ denotes the $j^{th}$ sampled point in the cell.
$f(;\bm{\theta})$ denotes the network with trainable parameters $\bm{\theta}$, and $\bm{z}_i$ denotes the latent shape-code for $\mathcal{C}_i$.
And $s(\hat{\bm{p}}_{i,j})$ here denotes the distance between the point $\hat{\bm{p}}_{i,j}$ and the surface $\mathcal{S}$ approximately represented by the discrete points $\mathcal{P}$, which can be calculated as
\begin{equation}
    s(\hat{\bm{p}}_{i,j}) = \|\hat{\bm{p}}_{i,j} - \bm{p}(\hat{\bm{p}}_{i,j})\|_2,
\end{equation}
where $\bm{p}(\hat{\bm{p}}_{i,j})=\arg\min_{\bm{p}\in\mathcal{S}}\|\bm{p}-\hat{\bm{p}}_{i,j}\|_2$, \ie, $\bm{p}$ is a point on the object surface $\mathcal{S}$.
For the convenience of calculation, we simply take the input point cloud $\mathcal{P}$ to substitute $\mathcal{S}$.

\paragraph{Geometric initialization for $\bm{\theta}$ -- }
We are learning a sign-agnostic implicit network via the unsigned distance, \ie, we only regress a absolute distance field as the level set.
This would lead to two kinds of solutions as mentioned in \cite{Atzmon2020_SAL}.
The first is a signed solution, where the points on both sides of the surface are predicted to be different signs.
And the other is a unsigned solution, whose predictions are of the same sign on both sides of the surface.
The signed one is the desired one, since it provides inside/outside indicator for the surface, which is a more feasible way to define the surface of $f(\bm{p},\bm{z}_i;\bm{\theta})=0$ with existing Marching Cubes algorithm \cite{}.
To make the trained network favor the signed solution, we follow the initialization of $\bm{\theta}$ proposed in \cite{Atzmon2020_SAL}, where a $R$-radius sphere is mimicked by the implicit function initially.
The difference is that we are handling local implicit function here, where the initialized local implicit function with specific $\bm{z}_i$ can be regarded as mimicking a local region on the $R$-radius sphere

\paragraph{Geometric transformation for $\bm{p}$ -- }
On the other hand, to accelerate the convergence of the network \jiabao{favor the signed solution} \yuxin{the signed solution is only encouraged by the initialization of $\bm{\theta}$ I guess?}, we also proposed geometric transformation for the points.
As mentioned in the earlier work \cite{Jiang2020}, the points within a cell should be firstly transformed into normalized local coordinates within the cell to $[-1, 1]$ to make the learning easier.
However, in our method, more than just normalizing the points, we also transform them into fitting a part of the $R$-radius sphere.
Formally, for a point $\bm{p}_{i,j}$ in the given input point set $\mathcal{P}_{\mathcal{C}_i}$ within cell $\mathcal{C}_i$, we perform the following transformation
\begin{equation}
    \begin{split}
        &\bm{p}_{i,j}^{\text{Normalized}} = \frac{2}{s_i}\cdot(\bm{p}_{i,j} -\bm{c}_i), \\
        &\bm{p}_{i,j}' = (\bm{p}_{i,j}^{\text{Normalized}} - \tilde{\bm{t}}_i )\cdot(\frac{R}{\tilde{r}_i}) ,
    \end{split}
\end{equation}
where $\bm{c}_i$ and $s_i$ are the center and scale of the $i^{th}$ cell respectively.
$\tilde{\bm{t}}_i$ and $\tilde{r}_i$ are the center and radius of a ball estimated by the all the normalized points in the cell via least square method.
Intuitively, we firstly normalize the points according to the cells, and then further transform them into part of the $R$-radius sphere our network initially mimics mentioned above.
Note that we perform estimation on $\tilde{\bm{t}}_i$ and $\tilde{r}_i$ iteratively, since the cells are adapting iteratively.

\subsubsection{Cells Adapting Strategy}
\label{sec:cells_adapting}
We allow two types of deformation to be performed on each cell during training, of which the first is that the center $\bm{c}$ of a cell is allowed to move from the initialized one, and the other is that the scale $\bm{s}$ of a cell is allowed to varies from the initialized one.
Apart from the above point-wise sign-agnostic fitting loss proposed to ensure individual cell could include a reasonable portion of the object as possible, we also proposed other principle for the deformations on cells -- all the input points should be overridden by the cells, cells should not be too large or too small, and neighbouring cells should be overlapping more or less.
To achieve these, we further proposed the \emph{covering loss}, the \emph{volume loss} and the rigid constraint of \emph{overlapping expansion}.

\paragraph{Covering loss -- }
We propose covering loss to make sure that the given input can be reconstructed completely, since two undesired but possible solutions for Eq.(\ref{eq:loss_SAF}) are to eliminate all the cells, or to move the cells to the spaces do not include any given points.
Suppose the union set of all the cells is $\bigcup{\mathcal{C}}$, covering loss is to encourage the input points to be overridden by the cells, which can be formally defined as,
\begin{equation}
    \mathcal{L}_{\text{Covering}} = \sum_{\bm{p}\not\in\cup{\mathcal{C}}} D_{\text{exterior}}(\bm{p}, \cup{\mathcal{C}}),
\end{equation}
where $D_{\text{exterior}}(\bm{p}, \cup{\mathcal{C}})$ denotes the distance between a point not covered by any cells and the cell being closest to it.
It's known as the box's exterior signed distance field, which can be defined as
\begin{equation}
    D_{\text{exterior}}(\bm{p}, \cup{\mathcal{C}}) = \min_{i\in\{1, \ldots, |\mathcal{C}|\}} \|\max\{|\bm{p}-\bm{c}_i|- \frac{\bm{s}_i}{2}, 0\}\|_2,
\end{equation}
where $\bm{s}_i=[s_i,s_i,s_i]^T\in\mathbb{R}^3$, where $\bm{c}_i$ and $s_i$ denotes the center and scale of the $i^{th}$ cell $\mathcal{C}_i$.
Also note that $\max$ and $|\cdot|$ here are element-wise operations.

\noindent\textbf{Volume loss -- }
To prevent unreasonable large scales of individual cells, we further proposed volume loss.
Formally, we penalize on the volume of each cell, as
\begin{equation}
    \mathcal{L}_{\text{Volume}} =  \sum_{i\in\{1, \ldots, |\mathcal{C}|\}} \max\{s_i, 0\},
\end{equation}
where $s_i$ denotes the scale of the $i^{th}$ cell $\mathcal{C}_i$.

\paragraph{Overlapping expansion -- }
Overlapping between different cells is desired for the post-processing in Section \ref{sec:post_processing} for two reasons .
The first is that we want to ensure the consistency of sign (interior/exterior) between the level sets predicted in different cells according to the overlapping part, and the second is that we want to smooth the junctions between different cells via interpolation.
To ensure overlapping between different cells, we further expand the scale of individual cell by a margin.
In other words, we perform a linear expansion on the scale of each cell $\mathcal{C}_i$, as
\begin{equation}
    s_{i}'=\alpha\cdot \max\{s_i, 0\}+ \beta,
\end{equation}
where $\alpha\geq1$ and $\beta\geq0$ are the hyper-parameters for the expansion.
$s_{i}'$ is the expanded scale used during inference process.
Though we can still not ensure adequate overlapping after this expansion, we have empirically proven that with appropriate $\alpha$ and $\beta$, this is enough to prevent unreasonable small scale of individual cells.

\subsubsection{The Combined Learning Objective}
We use the following combined objective to learn the local self-similarities with local implicit auto-decoder,
\begin{equation}
    \mathcal{L}(\mathcal{P}) = \mathcal{L}_{\text{SAF}}(\mathcal{P}) + \lambda_1\cdot\mathcal{L}_{\text{Covering}} + \lambda_2\cdot\mathcal{L}_{\text{Volume}}
\end{equation}
where $\lambda_1$ and $\lambda_2$ are penalty parameters.

\subsection{Post-processing of Local Surface}
\label{sec:post_processing}
In the previous section, local regions are learned without neighbouring spatial information, and thus, there might be error directly uniting local surfaces generated by the local implicit auto-decoder.
In this section, we proposed two post-processing strategies based on the overlapping part of different cells, including sign consistency and implicit fields interpolation to make the overall shape more reasonable.

\subsubsection{cells sign flipping}
\label{sec:sign_consi_flip}

\jiabao{Note that we borrow the idea from paper ``Surface reconstruction from unorganized points'', see section 3.3 ``Consistent Tangent Plane Orientation'' of this paper.} \yuxin{cite it in the appropriate position please.}
Local implicit auto-decoder is based on cells, so sign discontinuity would only appear between cells.
In the subsequent description, with a slight abuse of expression, we also use a cell to represent the local implicit auto-decoder based on the cell, which are self-clear in the context.
To make the internal and external information of the object globally consistent, we only need to perform \emph{cells sign flipping} to adjust the sign for each cell.
To be more specific, if the predictions between two adjacent cells on the object internal/external are discontinuous, we expect to flip the signs of the predictions generated by one of the two cells.
To achieve this goal, we leverage the strategies based on minimum spanning tree.
First, we need to construct a connected, edge-weighted undirected graph $G=(\mathcal{V}, \mathcal{E})$ that connects all the vertices together.
Regarding all the cells as vertices in the undirected graph, as $\mathcal{V} = \mathcal{C}$, we set an edge between two vertices if the cells are overlapped, as $\mathcal{E} = \{(\mathcal{V}_i,\mathcal{V}_j)|\mathcal{V}_i,\mathcal{V}_j\in \mathcal{V} \;\text{and}\; \mathcal{V}_i\bigcap \mathcal{V}_j \not= \emptyset \;\text{and}\; i\not=j \}$.
The weight $w_{\mathcal{E}_{i,j}}$ of an edge $\mathcal{E}_{i,j}$ between two vertices $\mathcal{V}_{i}$ and $\mathcal{V}_{j}$ is computed via 
\begin{equation}
    \begin{split}
        w_{\mathcal{E}_{i,j}} = &\min\{ \sum_{k\in\{1, \ldots, |\hat{\mathcal{P}}_{\mathcal{C}_i\cap \mathcal{C}_j}|\}} \big|f(\hat{\bm{p}}_{k}, \bm{z}_i;\bm{\theta}) - f(\hat{\bm{p}}_{k}, \bm{z}_j;\bm{\theta})\big|, \\
        &\sum_{k\in\{1, \ldots, |\hat{\mathcal{P}}_{\mathcal{C}_i\cap \mathcal{C}_j}|\}} \big|f(\hat{\bm{p}}_{k}, \bm{z}_i;\bm{\theta}) + f(\hat{\bm{p}}_{k}, \bm{z}_j;\bm{\theta})\big|\},        
    \end{split}
    \label{eq:tree_weight}
\end{equation}
\jiabao{maybe we should redefine the above formula by changing the $+$/$-$ to signs $+1$/$-1$ and putting them below the $\min$ operator for clarity?}
\yuxin{maybe? revise it directly please.}
where $\hat{\mathcal{P}}_{\mathcal{C}_i\cap \mathcal{C}_j}$ denotes the set of points uniformly sampled in the overlapping part of two cells, $\bm{z}_i$ and $\bm{z}_j$ denotes latent shape-code optimized for $\mathcal{C}_i$ and $\mathcal{C}_j$ respectively.
The weight would be chosen to be the former one (the signs of the cells staying the same) when there is no discontinuity between $\mathcal{C}_i$ and $\mathcal{C}_j$, while the latter one (one of the sign of the cell being flipped) when there is a discontinuity.

We additional add a ``sign'' attribute for each vertex, where $1$ indicates there is no sign flipping and $-1$ indicates there exists sign flipping.
Technologically, we follow Prim's algorithm \cite{} to greedily find a minimum spanning tree for the edge-weighted undirected graph we constructed, and endow the sign for each vertex during this process.
In more detail, we initialize $\mathcal{V}_{\text{Tree}}=\{\mathcal{V}_i\}$, where $\mathcal{V}_i$ is a random vertex in the edge-weighted undirected graph, and we initialize $\mathcal{E}_{\text{Tree}}=\emptyset$.
We set the sign of the first vertex we pick to be $1$.
To construct the minimum spanning tree, we take the minimum weighted-edge $\mathcal{E}_{u,v}$ in $\mathcal{E}$, where $\mathcal{V}_u\in\mathcal{V}_{\text{Tree}}$ and $\mathcal{V}_v\in\mathcal{V}\verb|\|\mathcal{V}_{\text{Tree}}$.
And we add $\mathcal{V}_v$ into $\mathcal{V}_{\text{Tree}}$ and add $\mathcal{E}_{u,v}$ into $\mathcal{E}_{\text{Tree}}$.
If the weighted of the edge takes the former one in Eq. (\ref{eq:tree_weight}), we set the sign of vertex $\mathcal{V}_u$ to be the same as that of vertex $\mathcal{V}_v$; and if the latter one is taken, the sign of vertex $\mathcal{V}_u$ is set to be different from vertex $\mathcal{V}_v$.
We repeat the above steps until $\mathcal{V}_{\text{Tree}}=\mathcal{V}$.
Once the tree is constructed, all vertices would have their sign properties.
And we perform sign flipping on each cell according to the sign property.

\subsubsection{Implicit Fields Interpolation}
Apart from sign discontinuity, there also might be numerical inconsistency between different cells.
To reduce such gap, we further proposed \emph{implicit fields interpolation}, which is to take the weighted average of the predictions in the overlapping part from different cells.
We normalize the weights according to the boxes' interior signed distance fields of the sampled point.
Given a sampled point $\hat{\bm{p}}_k$ falling in the overlapping part of a $m$ different cells $\{\mathcal{C}_{\hat{1}}, \ldots, \mathcal{C}_{\hat{m}}\}$ (the subscript of $\hat{i}$ here denotes a sorted order for the $i^{th}$ cell from these $m$ cells), the weight $w_{\hat{i}}$ can be defined as
\begin{equation}
    w_{\hat{i}}(\hat{\bm{p}}_k) = \frac{D_{\text{interior}}(\hat{\bm{p}}_k, \mathcal{C}_{\hat{i}})}{\sum_{j=\hat{1}}^{\hat{m}} D_{\text{interior}}(\hat{\bm{p}}_k, \mathcal{C}_{j})}
\end{equation}
where $D_{\text{interior}}$ denotes the box's interior signed distance filed, which can be defined as 
\begin{equation}
    D_{\text{interior}}(\hat{\bm{p}}_k, \mathcal{C}_{\hat{i}}) = \big|\min_{k=\{1,2,3\}}\{\pi_k(\frac{\bm{s}_{\hat{i}}}{2} - |\hat{\bm{p}}_k-\bm{c}_{\hat{i}}|)\}\big|,
\end{equation}
where $\bm{s}_{\hat{i}}=[s_{\hat{i}},s_{\hat{i}},s_{\hat{i}}]^T\in\mathbb{R}^3$, where $\bm{c}_{\hat{i}}$ and $s_{\hat{i}}$ denotes the center and scale of the ${\hat{i}}^{th}$ cell $\mathcal{C}_{\hat{i}}$.
$\pi_k(\cdot)$ is an operator selecting the $k^{th}$ element from the given input.
Also note that $\min$ and $|\cdot|$ here are element-wise operations, where $\min$ is taking the smallest value among all dimensions of the given vector.

With this, the the weighted value $f'(\cdot)$ of the sampled point $\hat{\bm{p}}_k$ is given as
\begin{equation}
    f'(\hat{\bm{p}}_k) = \sum_{j\in\{\hat{1},\ldots,\hat{m}\}} w_{j}(\hat{\bm{p}}_k) \cdot \text{sign}(\mathcal{C}_j) \cdot f(\hat{\bm{p}}_k, \bm{z}_j;\bm{\theta}),
    \label{eq:final_output}
\end{equation}
where $w_{j}(\hat{\bm{p}}_k)$ is the interpolation weight corresponding to the point $\hat{\bm{p}}_k$ in cell $\mathcal{C}_j$, $\text{sign}(\mathcal{C}_j)$ is the sign attribute generated for the cell $\mathcal{C}_j$ as mentioned in Section \ref{sec:sign_consi_flip}, and $f(\hat{\bm{p}}_k, \bm{z}_j;\bm{\theta})$ is the prediction from the local implicit network.

\subsubsection{Meshing}
With Eq.(\ref{eq:final_output}), we can use Marching Cubes \cite{} to extract the zero level-set to reconstruct a water-tight surface.

}

\vspace{-0.2cm}
\section{Experiments}
\label{sec:experiments}
\vspace{-0.1cm}

In this section, we present setups, comparative results, and robustness test to verify the efficacy of our proposed SAIL-S3, by comparing with the state-of-the-art methods for surface reconstruction from raw point clouds. \textbf{\emph{Ablation studies are given in the supplementary material.}}

\vspace{-0.1cm}
\vspace{0.1cm}\noindent\textbf{Datasets -- }We conduct experiments on the ShapeNet \cite{chang2015shapenet} and ThreeDScans \cite{threedscans} datasets that respectively contain synthetic objects and objects of real scans. For ShapeNet, we randomly select $100$ objects of \emph{chair}; for ThreeDScans, we randomly select $30$ sculptures. These objects are selected due to their complex shape topologies. In addition, we evaluate the robustness of our method by adding point-wise Gaussian noise to sculptures from ThreeDScans. We pre-process these object surfaces by centering their origins and scaling them uniformly to fit within the unit sphere. We then sample points at densities of $50,000$ and $100,000$ respectively from each instance of ShapeNet and ThreeDScans as the raw inputs. For comparative methods that require oriented surface normals, we compute the normals via tangent plane estimation \cite{hoppe1992surface} and reorient the directions via minimal spanning tree, which follows \cite{hoppe1992surface}.

\vspace{-0.1cm}
\vspace{0.1cm}\noindent\textbf{Implementation Details -- }
We adopt a $6$-layer MLP as our local implicit model, and initialize it according to Section \ref{sec:method_sign_agnostic}.
We initialize elements in the latent $\bm{z}$ by sampling from $\mathcal{N}(0, (1\times10^{-3})^2)$.
During learning, we optimize the objective (\ref{EqnLossOverall}) for $40,000$ iterations using Adam \cite{kingma2014adam}, with initial learning rates of $1\times{10}^{-3}$ for $\bm{\theta}$ and $\bm{z}$, and $3\times{10}^{-4}$ for subfield center $\bm{c}$ and scale $a$. The learning rates decay by $0.2$ at $20,000$, $30,000$, $35,000$ and $38,000$ iterations.
We set $\lambda_1 = 3\times 10^{-4}$, $\lambda_2 = 1.0$, $\lambda_3 = 1.0$ in the objective (\ref{EqnLossOverall}).
It takes around $70 \text{ms}$ per iteration, and better results require more iterations.

\vspace{-0.1cm}
\vspace{0.1cm}\noindent\textbf{Evaluation Metrics -- }
We randomly sample $100,000$ points respectively on the ground truth and reconstructed mesh, and use the metrics of Chamfer Distance (CD), Normal Consistency (NC) and F-score (F) to quantitatively evaluate different methods, where F is evaluated under the threshold of $0.005$.

\vspace{-0.1cm}
\vspace{0.1cm}\noindent\textbf{Comparative Methods -- }
We compare our method with three categories of existing methods, including the Screened Poisson Surface Reconstruction (SPSR) \cite{kazhdan2013screened}, global fitting methods such as Implicit Geometric Regularization (IGR) \cite{gropp2020implicit} (with normal data) and Sign Agnostic Learning (SAL) \cite{Atzmon2020_SAL} (without normal data), and locally learned methods such as Local implicit Grid (LIG) \cite{jiang2020local}, Convolutional Occupancy Networks (CON) \cite{Peng2020}, and Points2Surf (P2S) \cite{erler2020points2surf}.
We implement SPSR in MeshLab \cite{cignoni2008meshlab} with the default hyper-parameters.
For IGR and SAL, we directly fit the training points without learning from auxiliary data; for LIG, CON and P2S, we use the provided pre-trained models and the default settings from the original papers.
Note that global fitting methods require no auxiliary data, which is the same as our method. We summarize working conditions of different methods in Table \ref{table:compare_chara}.

\begin{table}
\begin{center}
\small
\scalebox{0.8}{
\begin{tabular}{|l|c|c|c|}
\hline
Methods & \makecell{No requirement on\\surface normals} & \makecell{No requirement on \\auxiliary data} & \makecell{Local \\ model} \\
\hline\hline
SPSR \cite{kazhdan2013screened} & $\times$  & $\checkmark$ & $\checkmark$  \\
\hline
IGR \cite{gropp2020implicit} & $\times$  & $\checkmark$  &  $\times$ \\
SAL \cite{Atzmon2020_SAL} & $\checkmark$ & $\checkmark$  &  $\times$ \\
\hline
LIG \cite{jiang2020local} &  $\times$ & $\times$  & $\checkmark$ \\
CON \cite{Peng2020} & $*$ &  $\times$  & $\checkmark$ \\
P2S \cite{erler2020points2surf} & $*$ & $\times$  & $\checkmark$ \\
\hline
SAIL-S3 & $\checkmark$ & $\checkmark$  & $\checkmark$ \\
\hline
\end{tabular}
}
\end{center}
\vspace{-0.5cm}
\caption{
Working condition summary of different methods.
Note that $*$ indicates that the method does not require surface normals during inference but does require during training.
Our proposed SAIL-S3 requires neither surface normals nor auxiliary training data.
}
\label{table:compare_chara}
\vspace{-0.6cm}
\end{table}

\nothing{
SPSR \cite{kazhdan2013screened}
IGR \cite{gropp2020implicit}
SAL \cite{Atzmon2020_SAL}
LIG \cite{jiang2020local}
ConvOccNet \cite{Peng2020}
Point2Surf \cite{erler2020points2surf}
}


\begin{figure*}[hptb]

        \centering
        \includegraphics[width=0.90\textwidth]{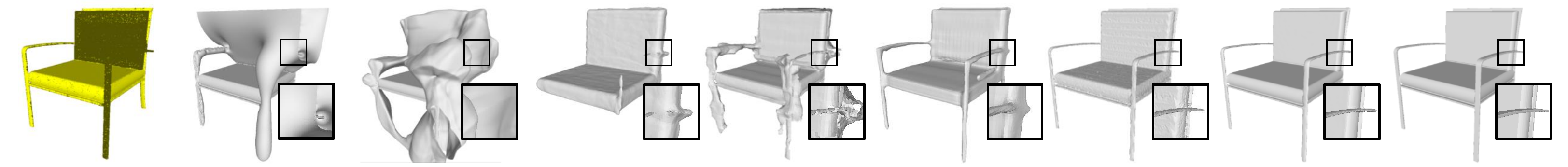} 

        \vspace{-0.05in}
        \centering
        \includegraphics[width=0.90\textwidth]{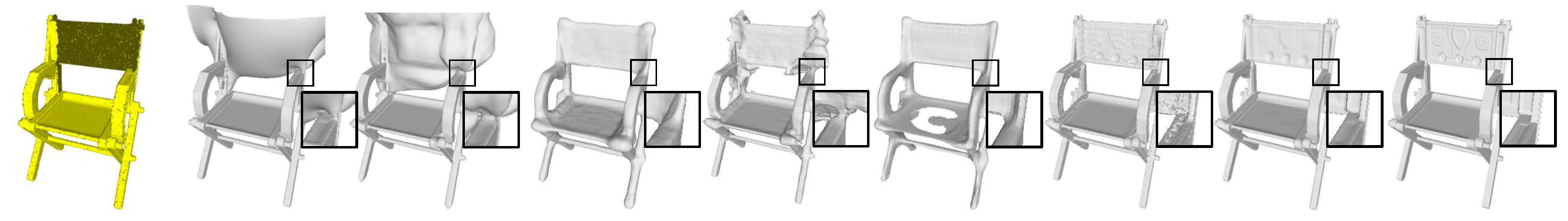} 

        \vspace{-0.1in}
        \centering
        \includegraphics[width=0.90\textwidth]{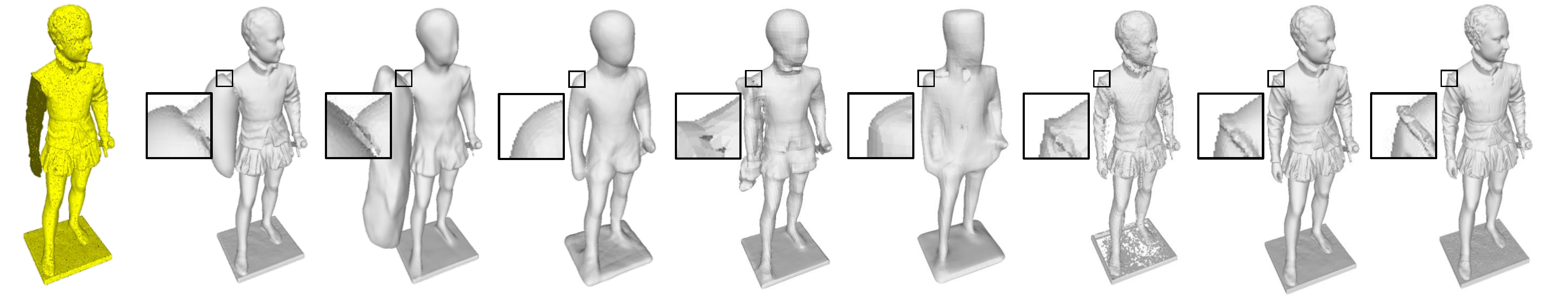} 

        \vspace{-0.05in}
        \centering
        \includegraphics[width=0.90\textwidth]{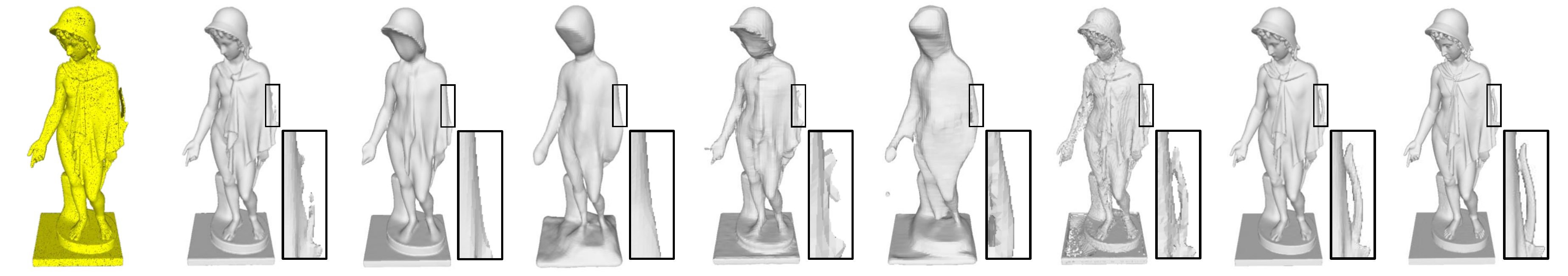} 

        \vspace{-0.05in}
        \centering
        \includegraphics[width=0.90\textwidth]{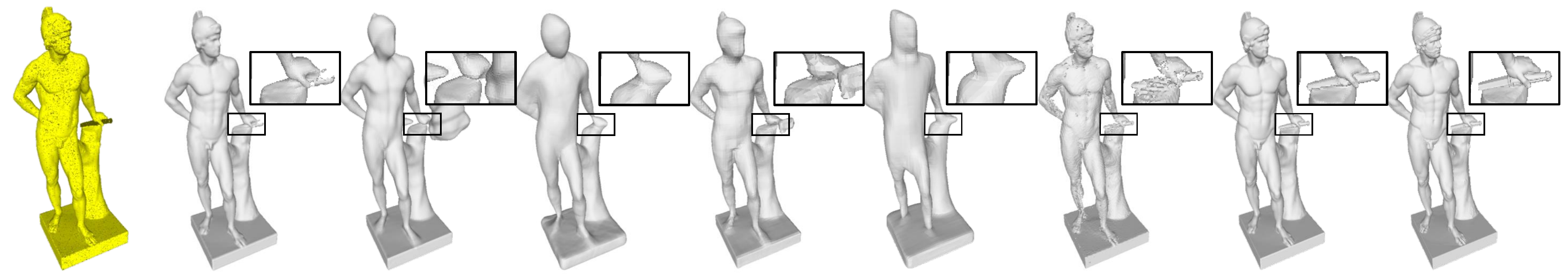} 

        \vspace{-0.05in}
        \scalebox{0.90}{
            \begin{tabular}{p{45pt}p{45pt}p{45pt}p{45pt}p{45pt}p{45pt}p{45pt}p{45pt}p{45pt}}
            \small Input PC & \
            \small SPSR \cite{kazhdan2013screened} & \
            \small IGR \cite{gropp2020implicit} &  \
            \small SAL \cite{Atzmon2020_SAL}  & \
            \small LIG \cite{jiang2020local}  & \
            \small CON \cite{Peng2020} &  \
            \small P2S \cite{erler2020points2surf} &  \
            \small SAIL-S3  & \
            \small GT\\
            \end{tabular}
        }

    \vspace{-0.2cm}
    \caption{
    Qualitative results of different methods on \emph{chair} instances in ShapeNet \cite{chang2015shapenet} (top two rows) and sculptures in ThreeDScans \cite{threedscans} (bottom three rows). Black points on the five inputs denote incorrect estimations of normal orientations.
    Note that IGR and SAL belong to global fitting methods, and LIG, CON, and P2S belong to locally learned methods. Refer to the supplementary material for more qualitative results.
    }
    \label{fig:exp_comp}
\end{figure*}
\vspace{-0.2cm}

\vspace{0.08cm}
\subsection{Comparative Results} \deadline{11.7}
\label{sec:exp_com}
\vspace{-0.08cm}

To demonstrate the efficacy of our method to reconstruct high-fidelity surfaces from raw point clouds, we conduct surface reconstruction experiments on \emph{chair} objects in ShapeNet \cite{chang2015shapenet} and sculptures in ThreeDScans \cite{threedscans}.
Qualitative results are shown in Figure \ref{fig:exp_comp}.
For the methods relying on global representations, such as IGR \cite{gropp2020implicit} and SAL \cite{Atzmon2020_SAL}, they fail to generalize to the complex topologies.
Conv. OccNet (CON) \cite{Peng2020} and Point2Surf (P2S) \cite{erler2020points2surf} perform much better, almost recovering the topologies of different chairs and sculptures; however, their details are either missing or rugged.
Results from those methods requiring accurate surface normals, such as SPSR \cite{kazhdan2013screened}, IGR \cite{gropp2020implicit}, and LIG \cite{jiang2020local}, have unpredictable errors including non-watertight or wrongly folded meshes, since estimated normals might not be accurate enough. 
In contrast, our proposed SAIL-S3 recovers both the correct topologies and surface details from the raw, un-oriented input points, without requiring any auxiliary data.
Quantitative results in Table \ref{table:exp_syn_recon} further confirm the superiority of our method over existing ones.


\begin{table}[htbp]
\begin{center}
    \vspace{-0.2cm}
    \scalebox{0.8}{
        \begin{tabular}{|l|c|c|c||c|c|c|}
        \hline
        Datasets    & \multicolumn{3}{c||}{ShapeNet\cite{chang2015shapenet}} & \multicolumn{3}{c|}{ThreeDScans \cite{threedscans}} \\
        \hline
        Methods            & CD $\downarrow$ & NC $\uparrow$ & F $\uparrow$  &       CD $\downarrow$ & NC $\uparrow$ & F $\uparrow$ \\
        \hline
        \hline
        SPSR \cite{kazhdan2013screened}         & 0.009 & 0.966 & 0.832   & \textbf{0.003} & 0.968 & 0.864 \\
        \hline
        IGR \cite{gropp2020implicit}          & 0.011 & 0.955 & 0.774     & 0.007 & 0.942 & 0.773   \\
        SAL \cite{Atzmon2020_SAL}          & 0.015 & 0.897 & 0.360        & 0.009 & 0.899 & 0.332 \\
        \hline
        LIG \cite{jiang2020local}          & 0.006 & 0.940 & 0.756        & 0.005 & 0.920 & 0.727 \\
        CON \cite{Peng2020}   & 0.011 & 0.876 & 0.269                     & 0.010 & 0.852 & 0.258 \\
        P2S \cite{erler2020points2surf}   & \textbf{0.003} & 0.928 & 0.798         & 0.005 & 0.869 & 0.669\\
        \hline
        SAIL-S3     & \textbf{0.003} & \textbf{0.981} & \textbf{0.884}    & \textbf{0.003} & \textbf{0.972} & \textbf{0.871} \\
        \hline
        \end{tabular}
    }
    \vspace{-0.2cm}
    \caption{Quantitative results of \emph{chair} instances in ShapeNet \cite{chang2015shapenet} and sculptures in ThreeDScans \cite{threedscans}. For CD, the smaller, the better; for NC and F, the larger, the better.}
    \label{table:exp_syn_recon}
\end{center}
\vspace{-0.4cm}
\end{table}

\begin{figure*}[hptb]
    \vspace{-0.05in}
    \centering
    \includegraphics[width=0.89\textwidth]{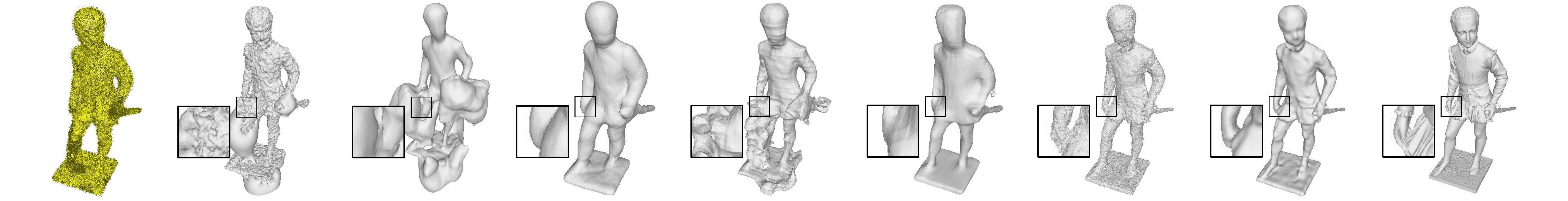}

    \vspace{-0.05in}
    \centering
    \includegraphics[width=0.89\textwidth]{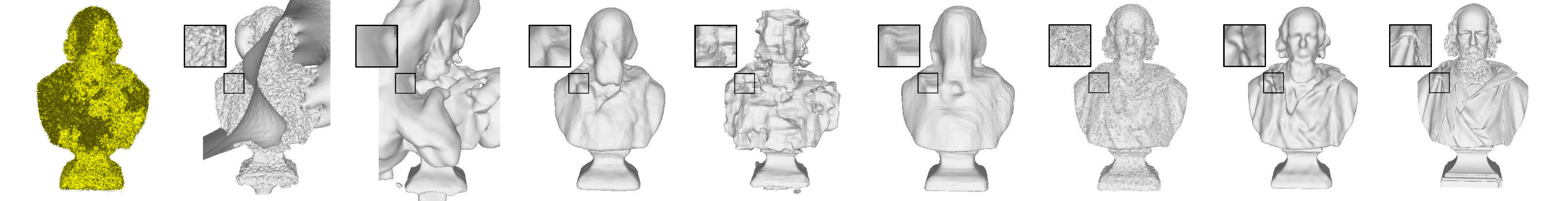}

    \vspace{-0.05in}
    \centering
    \includegraphics[width=0.89\textwidth]{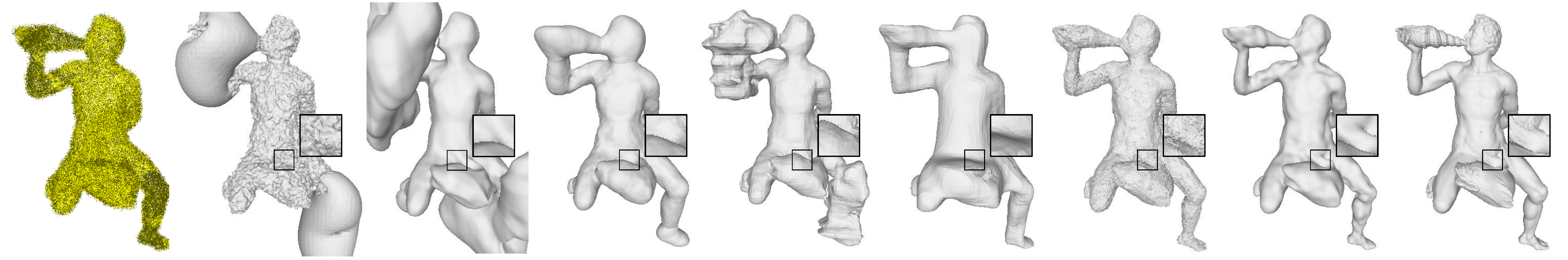}

    \vspace{-0.05in}
        \scalebox{0.89}{
        \begin{tabular}{p{45pt}p{45pt}p{45pt}p{45pt}p{45pt}p{45pt}p{45pt}p{45pt}p{45pt}}
        \small Input PC & \
        \small SPSR \cite{kazhdan2013screened} & \
        \small IGR \cite{gropp2020implicit} &  \
        \small SAL \cite{Atzmon2020_SAL}  & \
        \small LIG \cite{jiang2020local}  & \
        \small CON \cite{Peng2020} &  \
        \small P2S \cite{erler2020points2surf} &  \
        \small SAIL-S3  & \
        \small GT\\
        \end{tabular}
    }
    \vspace{-0.2cm}
    \caption{
    Qualitative results of different methods when adding point-wise Gaussian noise of standard deviation 0.01 to input points of sculptures in ThreeDScans \cite{threedscans}. Black points on the three inputs denote incorrect estimations of normal orientations.
    Note that IGR and SAL belong to global fitting methods, and LIG, CON, and P2S belong to locally learned methods.
    }
    \label{fig:robustness}
\end{figure*}
\vspace{-0.2cm}

\vspace{-0.3cm}
\subsection{Robustness Evaluation}
\label{sec:exp_rob}
\vspace{-0.1cm}
We further evaluate the robustness of different methods against noisy inputs. We use sculptures from ThreeDScans \cite{threedscans}, and add point-wise Gaussian noise of varying levels.
As shown in Figure \ref{fig:robustness}, adding noise to the input points indeed degrades the performance of different methods, particularly for those methods that require estimation of surface normals. 
Our proposed SAIL-S3 stays more robust against the noise. Quantitative comparisons with different noise levels are given in the supplementary material.


\vspace{-0.3cm}
\section{Acknowledgement}
\vspace{-0.1cm}
This work was supported in part by the National Natural Science Foundation of China (Grant No.: 61771201), the Program for Guangdong Introducing Innovative and Entrepreneurial Teams (Grant No.: 2017ZT07X183), and the Guangdong R\&D key project of China (Grant No.: 2019B010155001).



{\small
	\bibliographystyle{ieee_fullname}
	\bibliography{paper}
}

\newpage
\appendix

\end{document}


\title{Sign-Agnostic Implicit Learning of Surface Self-Similarities for Shape Modeling and Reconstruction from Raw Point Clouds\\
	-- \textit{Supplemental Material} --}

\maketitle
\pagestyle{empty}  
\thispagestyle{empty} 


\section{Least Squares Fitting of A Sphere to Observed Points}
We introduce how to estimate the center $\bar{\bm{t}}$ and the radius $\bar{\bm{r}}$ given the observed points $\{{\bm{p}}\}$ using least squares, by solving the following problem. Its solution can be easily obtained in a closed form.
\begin{equation}\label{EqnLeastSquareFitting}
    \begin{split}
        & \qquad (\bar{r}, \bar{\bm{t}}) = \arg\min_{\bar{r}', \bar{\bm{t}}'} \|\bm{A}\bm{b}-\bm{y}\|_2, \\
        &\bm{y}=\left[
    \begin{matrix}
        \| \bm{p}^1 \|^2_2 \\
        \vdots \\
        \| \bm{p}^n \|^2_2 \\
     \end{matrix}
     \right],
    \bm{b}=\left[
    \begin{matrix}
        \bar{t}_1'\\
        \bar{t}_2'\\
        \bar{t}_3'\\
        \bar{r}'^2 - \|\bar{\bm{t}'}\|_2^2
    \end{matrix}
     \right],\\
    &\bm{A}=\left[
        \begin{matrix}
            2{p}_1^{1} & 2{p}_2^{1} & 2{p}_3^{1} & 1 \\
            \vdots & \vdots & \vdots & \vdots \\
            2{p}_1^{n} & 2{p}_2^{n} & 2{p}_3^{n} & 1 \\
         \end{matrix}
         \right].
    \end{split}
\end{equation}
Its closed-form solution is given by
\begin{equation}
    \begin{split}
        \bar{r} &= \sqrt{{b}_{4} + \| \bm{b}_{[1:3]} \|_2^2}, \\
        \bar{\bm{t}} &= \bm{b}_{[1:3]},
    \end{split}
\end{equation}
where $\bm{b}$ can be computed by $ (\bm{A}^T \bm{A})^{-1} \bm{A}^T \bm{y}$.

\section{Minimum Spanning Tree for a Global Post-Optimization of Local Sign Flipping}
I explain here how we have used minimum spanning tree (MST) in Section 4.3 for a global post-optimization of local sign flipping. 
Simply put, MST grows a set $\mathcal{V}_{\textrm{MST}}$ as a tree by selecting the vertices from $\mathcal{V}$. Starting from $\mathcal{V}_{\textrm{MST}} = \emptyset$, MST randomly selects a vertex from $\mathcal{V}$, denoted as $v_1$, and assigns $h(v_1) = 1$.
The tree then grows iteratively by selecting from $\mathcal{V} / \mathcal{V}_{\textrm{MST}}$ the vertex that has the lowest edge weight in $\mathcal{W}$ to connect with any vertex in $\mathcal{V}_{\textrm{MST}}$, and ensures not to form a closed loop with those previous selected edges.
In any iteration, denote the selected edge as $e_{i, j}$, $v_j \in \mathcal{V} / \mathcal{V}_{\textrm{MST}}$ is the selected vertex and $v_i$ is the corresponding vertex already in $\mathcal{V}_{\textrm{MST}}$; we set $h(v_j) = h(v_i)$ when $w_{i, j}^1(e_{i,j}) < w_{i, j}^0(e_{i,j})$, and $h(v_j) = - h(v_i)$ otherwise.
The tree is spanned until $|\mathcal{V}_{\textrm{MST}}| = |\mathcal{V}|$, and we have the signs $\{ h(v_i) \}_{i=1}^N$ determined for all the vertices by then.


\section{Proof of Corollary 4.1}
\noindent\textbf{Corollary 4.1.}
\textit{
Let $f:\mathbb{R}^{3+d}\to\mathbb{R}$ be an $l$-layer MLP with ReLU activation $\nu$.
That is, $f(\bm{p},\bm{z})=\bm{w}^T\nu(\bm{W}^{l}(\cdots\nu(\bm{W}_{\bm{p}}^{1}\bm{p}+\bm{W}_{\bm{z}}^{1}\bm{z}+\bm{b}^{1}))+\bm{b}^{l})+c$, where $\bm{W}_{\bm{p}}^{1}\in\mathbb{R}^{d^{1}_{\textrm{out}}\times 3}$ and $\bm{W}_{\bm{z}}^{1}\in\mathbb{R}^{d^{1}_{\textrm{out}}\times d}$ denote the weight matrices of the first layer, and $\bm{b}^{1}\in\mathbb{R}^{d^{1}_{\textrm{out}}}$ denotes the bias; $\bm{W}^{i}\in\mathbb{R}^{d^{i}_{\textrm{out}}\times d^{i-1}_{\textrm{out}}}$ and $\bm{b}^{i}\in\mathbb{R}^{d^{i}_{\textrm{out}}}$ denote parameters of the $i^{th}$ layer; $\bm{w}\in\mathbb{R}^{d_{\textrm{out}}^{l}}$ and $c\in\mathbb{R}$ are parameters of the last layer; $\bm{p}\in\mathbb{R}^3$ is the input point, and $\bm{z}\in\mathbb{R}^{d}$ is the latent code, whose elements follow the i.i.d. normal $\mathcal{N}(0,\sigma_{z}^2)$.
Let $\bm{w}=\sqrt{\frac{\pi}{d_{\textrm{out}}^{l}}}\bm{1}$, $c=-\bar{r}$, $\bar{r}>0$, let all entries of $\bm{W}^{i}$ ($2\leq i\leq l$) follow i.i.d. normal $\mathcal{N}(0,\frac{2}{d_{\textrm{out}}^{i}})$, let entries of $\bm{W}_{\bm{p}}^{1}$ follow i.i.d. normal $\mathcal{N}(0,\frac{2}{d_{\textrm{out}}^{1}})$, and let $\bm{b}^i=\bm{0}$ ($2\leq i\leq l$).
If $\bm{W}_{\bm{z}}^{1}=\bm{W}_{\bm{p}}^{1}[\bm{I}\in\mathbb{R}^{3\times 3}, \bm{0}\in\mathbb{R}^{3\times {(d-3)}}]$ and $\bm{b}^1=-\bm{W}_{\bm{p}}^{1}\bar{\bm{t}}$, then $\lim_{\sigma_z \to 0} f(\bm{p},\bm{z}) = \|\bm{p}-\bar{\bm{t}}\| - \bar{r}$.
That is, $f$ is approximately the signed distance function to a 3D sphere of radius $\bar{r}$ centered at $\bar{\bm{t}}$.
}

\begin{proof}
To prove this theorem, we reduce the problem to a single hidden layer network.
By plugging $\bm{W}_{\bm{z}}^1$, $\bm{b}^1$ in $f$ we get $f(\bm{p},\bm{z})=\bm{w}^T\nu(\bm{W}^1_{\bm{p}}(\bm{p}+\bm{z}_{[1:3]}-\bar{\bm{t}}))+c$.
Let $\bm{x}\in\mathbb{R}^3 = \bm{p}+\bm{z}_{[1:3]}-\bar{\bm{t}}$ and further plug $\bm{w}$, $c$ in $f$, we get $f(\bm{x}) = \sqrt{\frac{\pi}{d_{\textrm{out}}^1}}\sum_{i=1}^{d_{\textrm{out}}^{1}}\nu(\bm{w}^1_{i,:}\cdot\bm{x})-\bar{r}$, where $\bm{w}^1_{i,:}$ is the $i^{th}$ row of $\bm{W}^1_{\bm{p}}$.
Let $\mu$ denotes the density of multivariate normal distribution $\mathcal{N}(0, \frac{2}{d_{\textrm{out}}^{1}} \bm{I}_{d_{\textrm{out}}^1})$.
The first term of $f(\bm{x})$ converges to $\|\bm{x}\|$, which is a direct consequence following Theorem $2$ in \cite{Atzmon2020_SAL}.
In other words,  $f(\bm{p}, \bm{z}) \approx \|\bm{p}-\bar{\bm{t}}+\bm{z}_{[1:3]}\| - \bar{r} \approx \|\bm{p}-\bar{\bm{t}}\| - \bar{r}$ when $\sigma_{\bm{z}} \to 0$.
Note that to make the assumption of $\bm{b}^1$ and $\bm{W}_{\bm{z}}^1$ to be true, $d_{\textrm{out}}^{1}$ should satisfy $d_{\textrm{out}}^{1} \geq 3$.
\end{proof}


\section{Ablation Studies}

\begin{figure}[hptb]
    \centerline{\includegraphics[width=0.49\textwidth]{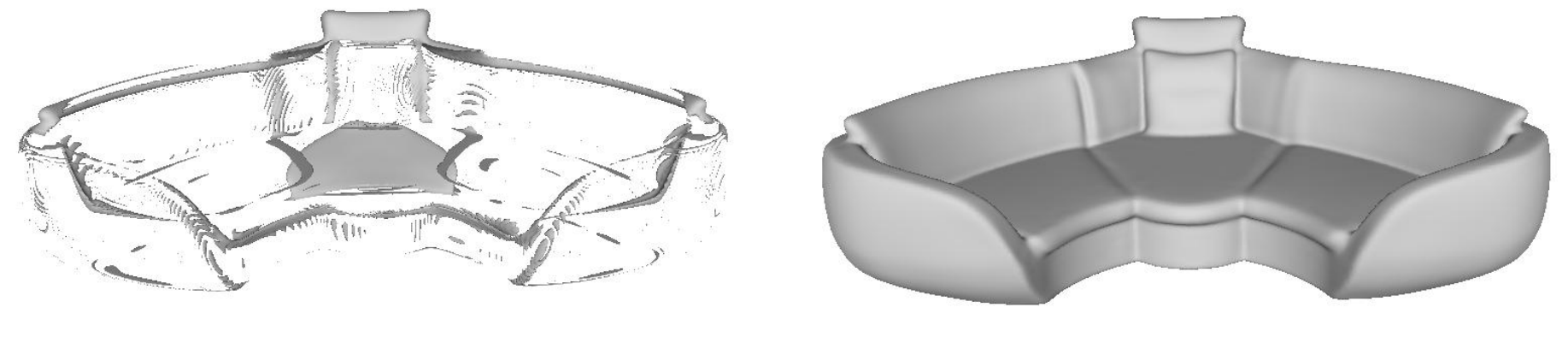}}
    \begin{tabular}{p{130pt}p{130pt}}
    \ W/O proper initialization & \quad \; SAIL-S3  \\
    \end{tabular}\vspace{-0.1cm}
    \caption{
    Without proper initialization, the learned local implicit function outputs $+/-$ boundaries only at isolated field regions (i.e. grey stripes in the left figure); note that only these boundaries can be extracted as surfaces via Marching Cubes \cite{lorensen1987marching}.
    }
    \label{fig:abla_init}
    \vspace{-0.1cm}
\end{figure}

\noindent\textbf{Initialization for Signed Solutions -- }
To evaluate the advantages of our proposed initialization for signed solution presented in Section 4.2.2, we conduct experiments that optimize SAIL-S3 with or without using the proposed initialization.
Figure \ref{fig:abla_init} shows that replacing our geometric initialization with random initialization will
generate results of isolated surface stripes, which are extracted from the incorrect signed solutions of implicit fields. 

\begin{figure}[hptb]
    \centerline{\includegraphics[width=0.49\textwidth]{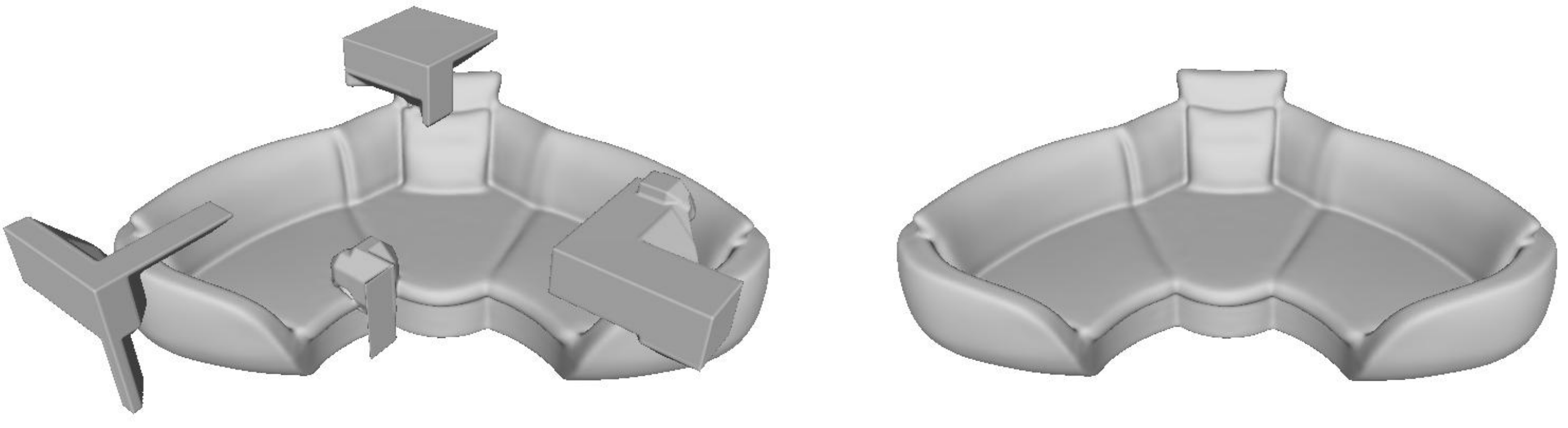}}
    \begin{tabular}{p{130pt}p{130pt}}
    \ W/O proper sign flipping & \quad \; SAIL-S3  \\
    \end{tabular}\vspace{-0.1cm}
    \caption{
    Without proper sign flipping, the resulting mesh may have blocky defects.
    }
    \label{fig:abla_sign_flip}
    \vspace{-0.1cm}
\end{figure}

\noindent\textbf{Local Sign Flipping -- }
Fig. \ref{fig:abla_sign_flip} shows that switching off the local sign flipping presented in Section $4.3$ would make the signs of different subfields inconsistent, ultimately resulting in surfaces with artifacts. 

\begin{figure}[hptb]
    \centerline{\includegraphics[width=0.49\textwidth]{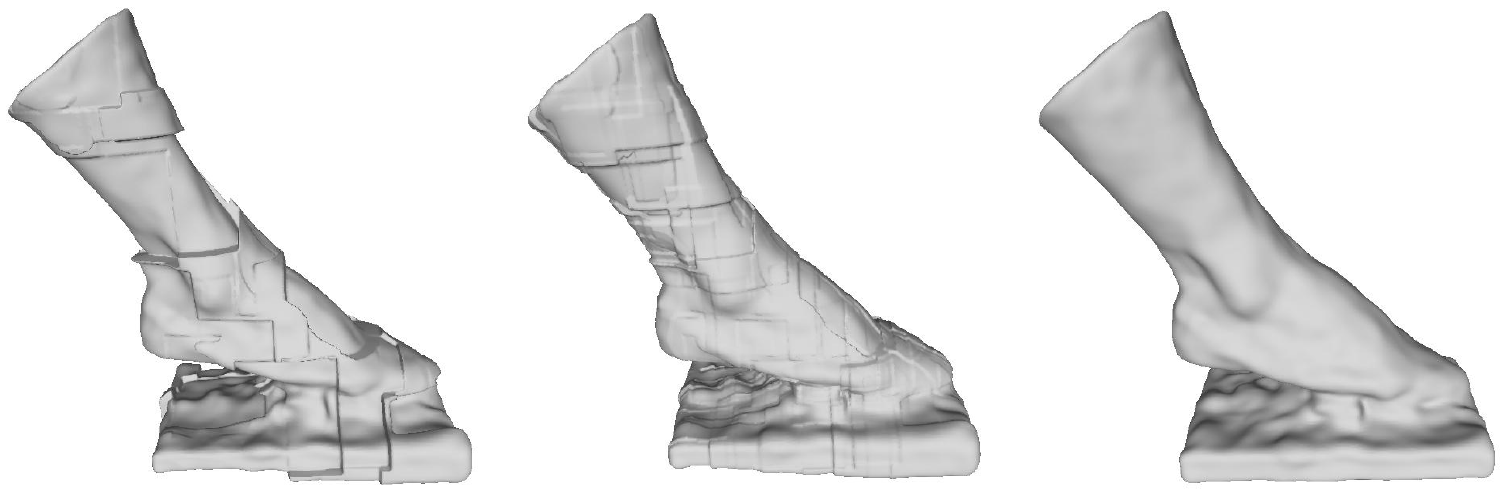}}
    \begin{tabular}{p{70pt}p{70pt}p{70pt}}
    \qquad \ \ Max & \qquad \ Average & \qquad SAIL-S3  \\
    \end{tabular}\vspace{-0.1cm}
    \caption{
    Example results of ablation study among interpolation by max pooling, interpolation by average pooling and our weighted interpolation.
    }
    \label{fig:abla_inter}
    \vspace{-0.1cm}
\end{figure}

\noindent\textbf{Interpolation of Local Fields --}
The validity of the proposed interpolation of local subfields can be checked by substituting with interpolation by max pooling or average pooling, as shown in Fig. \ref{fig:abla_inter}.

\section{Additional Qualitative Results } \label{app:quality}
\begin{figure*}[hptb]
    \centering
    \includegraphics[width=0.99\textwidth]{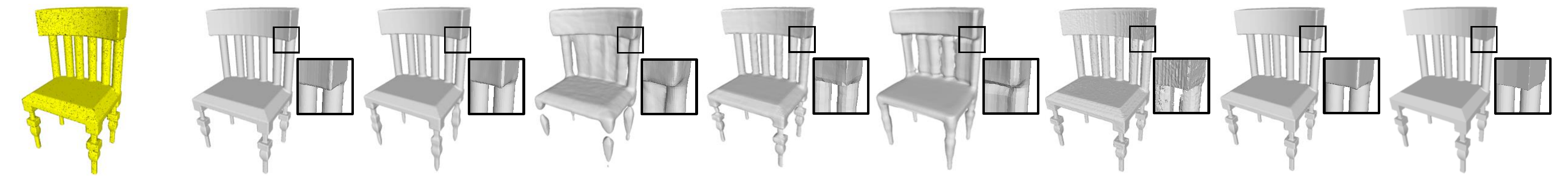}\vspace{0.5cm}\\
    \centering
    \includegraphics[width=0.99\textwidth]{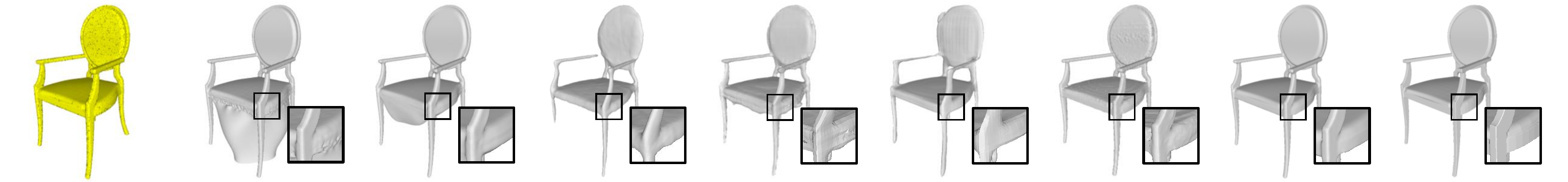}\vspace{0.5cm}\\
    \centering
    \includegraphics[width=0.99\textwidth]{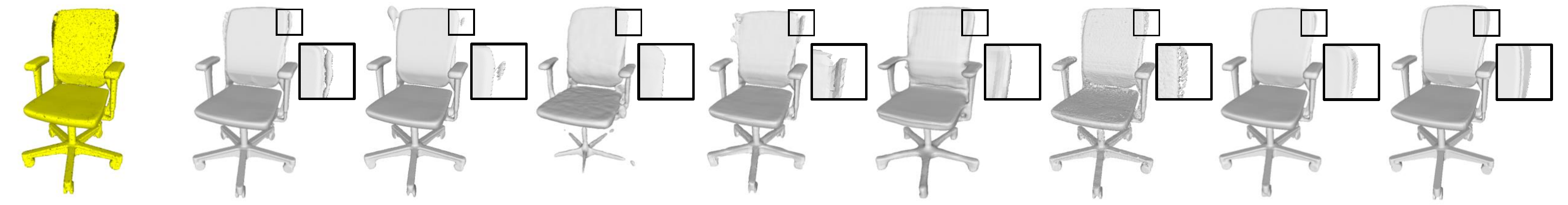}\vspace{0.5cm}\\
    \centering
    \includegraphics[width=0.99\textwidth]{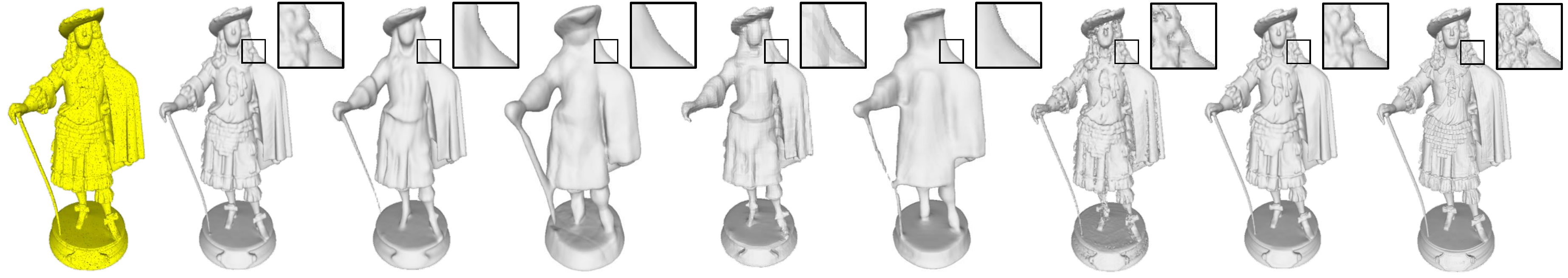}\vspace{0.5cm}\\
    \centering
    \includegraphics[width=0.99\textwidth]{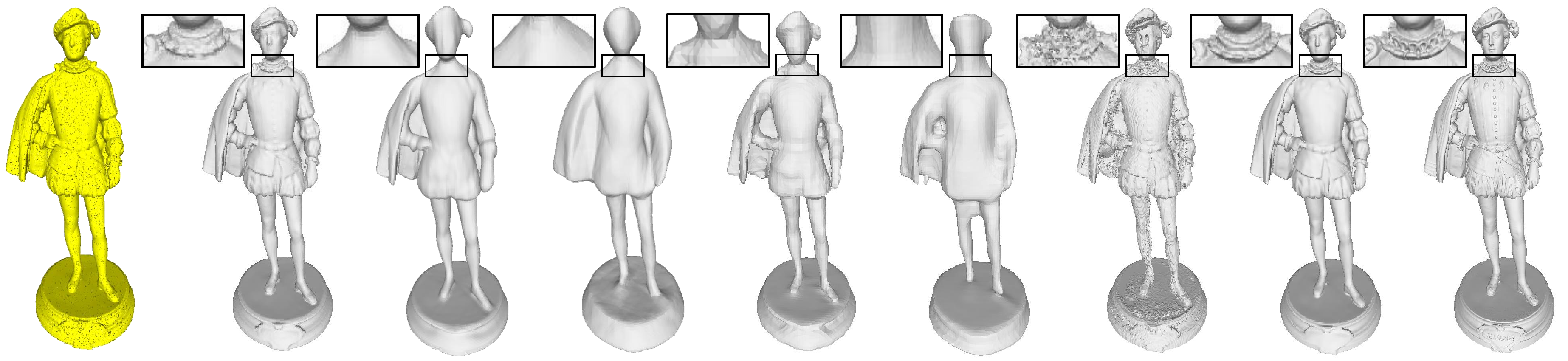}\vspace{0.5cm}\\
    \begin{tabular}{p{45pt}p{45pt}p{45pt}p{45pt}p{45pt}p{45pt}p{45pt}p{45pt}p{45pt}}
    \small \ \ \ Input PC & \ 
    \small \ \ \ SPSR \cite{kazhdan2013screened} & \
    \small \ \ \ IGR \cite{gropp2020implicit} &  \
    \small \ \ \ SAL \cite{Atzmon2020_SAL}  & \ 
    \small \ \ LIG \cite{jiang2020local}  & \
    \small CON \cite{Peng2020} &  \
    \small \ P2S \cite{erler2020points2surf} &  \
    \small SAIL-S3  & \
    \small \ GT\\
    \end{tabular}
    \caption{
    Additional qualitative reconstruction results of ShapeNet \cite{chang2015shapenet} (top three rows) and ThreeDScans \cite{threedscans} (bottom two rows), black regions of the input point cloud denote incorrect normal orientation.
    Note that IGR \cite{gropp2020implicit} and SAL \cite{Atzmon2020_SAL} belong to global fitting methods; while LIG \cite{jiang2020local}, CON \cite{Peng2020} and P2S \cite{erler2020points2surf} belong to the local learning methods.
    Lens are used to highlight the differences among the comparative methods.
    }
    \label{fig:exp_comp_supp}
\end{figure*}

\begin{figure*}[hptb]
    \centering
    \includegraphics[width=0.95\textwidth]{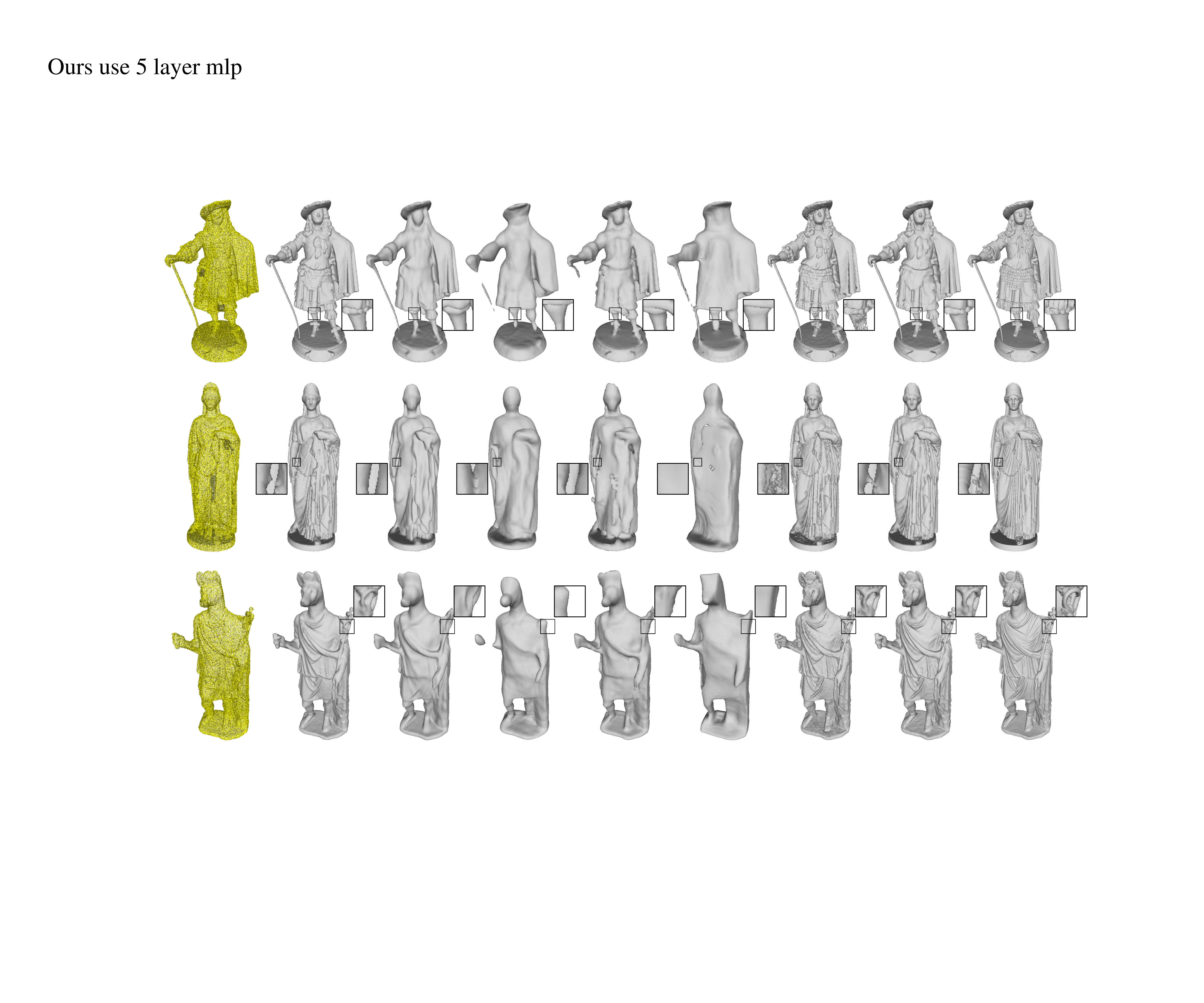}
    \begin{tabular}{p{50pt}p{45pt}p{45pt}p{45pt}p{35pt}p{35pt}p{45pt}p{45pt}p{45pt}}
    \small \ \ \ \ \ \ \ \ \ Input PC &
    \small \ \ \ SPSR \cite{kazhdan2013screened} &
    \small \ \ IGR \cite{gropp2020implicit} & 
    \small SAL \cite{Atzmon2020_SAL}  & 
    \small LIG \cite{jiang2020local}  &
    \small CON \cite{Peng2020} & 
    \small \ \ P2S \cite{erler2020points2surf} & 
    \small SAIL-S3  &
    \small GT\\
    \end{tabular}\vspace{-0.3cm}
    \caption{
    Qualitative results of different methods when adding point-wise Gaussian noise of standard deviation 0.001 to input points of sculptures in ThreeDScans \cite{threedscans}. Black points on the three inputs denote incorrect estimations of normal orientations.
    Note that IGR and SAL belong to global fitting methods, and LIG, CON, and P2S belong to locally learned methods.
    }
    \label{fig:noisy_0.001}
\end{figure*}

\begin{figure*}[hptb]
    \centering
    \includegraphics[width=0.95\textwidth]{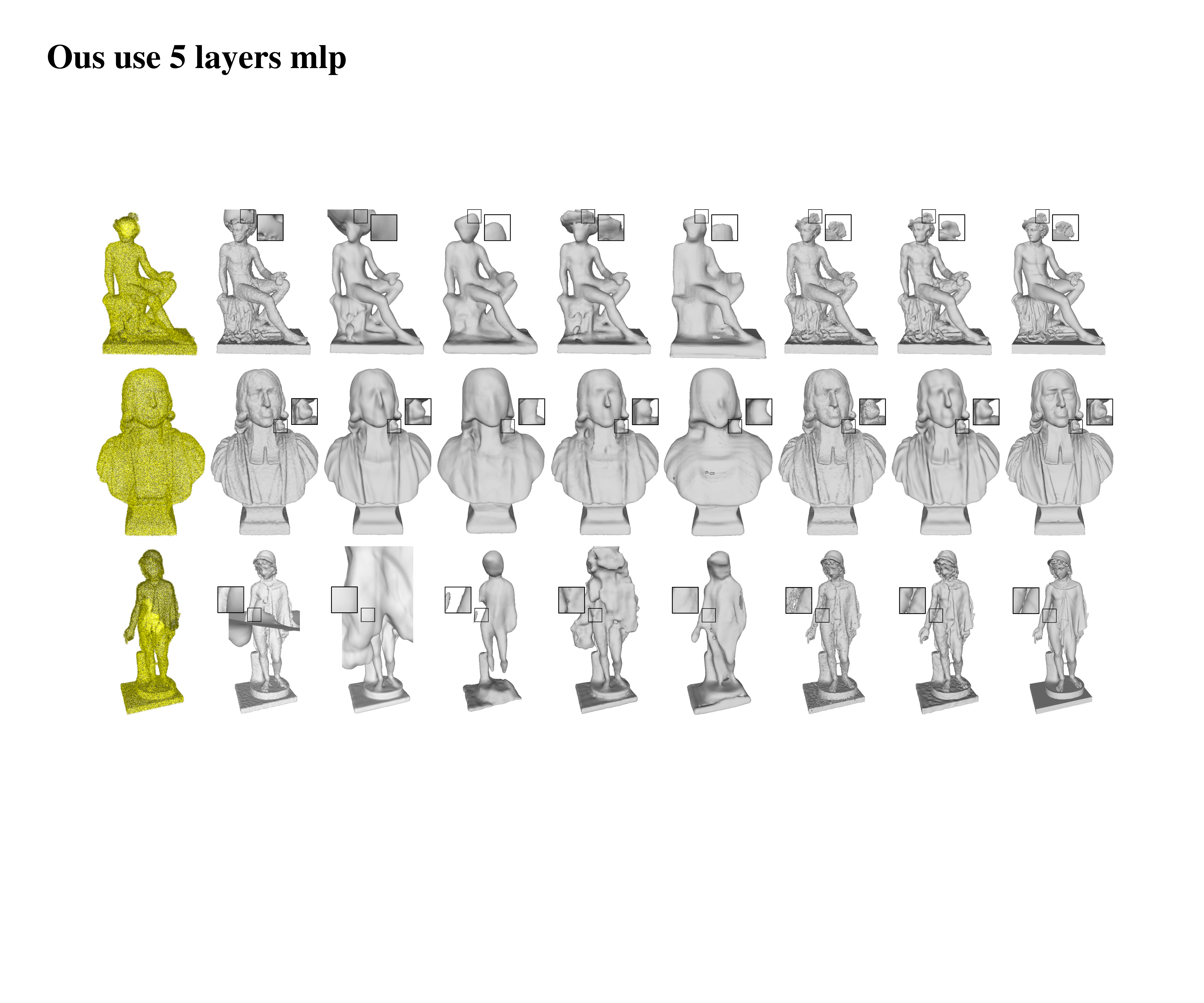}
    \begin{tabular}{p{50pt}p{45pt}p{45pt}p{45pt}p{35pt}p{35pt}p{45pt}p{45pt}p{45pt}}
    \small \ \ \ \ \ \ \ \ \ Input PC &
    \small \ \ \ SPSR \cite{kazhdan2013screened} &
    \small \ \ IGR \cite{gropp2020implicit} & 
    \small SAL \cite{Atzmon2020_SAL}  & 
    \small LIG \cite{jiang2020local}  &
    \small CON \cite{Peng2020} & 
    \small \ \ \ \ P2S \cite{erler2020points2surf} & 
    \small SAIL-S3  &
    \small \ \ GT\\
    \end{tabular}\vspace{-0.3cm}
    \caption{
    Qualitative results of different methods when adding point-wise Gaussian noise of standard deviation 0.005 to input points of sculptures in ThreeDScans \cite{threedscans}. Black points on the three inputs denote incorrect estimations of normal orientations.
    Note that IGR and SAL belong to global fitting methods, and LIG, CON, and P2S belong to locally learned methods.
    }
    \label{fig:noisy_0.005}
\end{figure*}

More qualitative results are given in Figure \ref{fig:exp_comp_supp}, Figure \ref{fig:noisy_0.001} and Figure \ref{fig:noisy_0.005}.
Our results are better than existing ones in terms of recovering smoother surfaces with more details. 

\section{Quantitative Results for Robustness Evaluation}
Table \ref{table:exp_noisy_recon_cd} shows the quantitative results for our robustness evaluation presented in Section 5.2. 
The CD results confirm that our method is more robust against noisy observations. 
\begin{table}[htbp]
\begin{center}
    \scalebox{1.0}{
        \begin{tabular}{|l|c|c|c|}
        \hline
        Methods            & CD@0.01 $\downarrow$ & CD@0.005 $\downarrow$ & CD@0.001 $\downarrow$   \\
        \hline
        \hline
        SPSR \cite{kazhdan2013screened} & 0.023 & 0.020  & $\bm{0.003}$   \\
        \hline
        IGR \cite{gropp2020implicit}   & 0.035 & 0.018 & 0.008    \\
        SAL \cite{Atzmon2020_SAL}     &  0.014 & 0.012 & 0.010  \\
        \hline
        LIG \cite{jiang2020local}     &  0.013 &  0.008 &  0.007   \\
        CON \cite{Peng2020}   & 0.016 & 0.014  & 0.011  \\
        P2S \cite{erler2020points2surf} & 0.008 & 0.008 & 0.005  \\
        \hline
        SAIL-S3 &  $\bm{0.004}$ & $\bm{0.003}$ & $\bm{0.003}$  \\
        \hline
        \end{tabular}
    }
    \caption{
    Quantitative results for noisy point clouds of sculptures in ThreeDScans \cite{threedscans}. For CD, the smaller, the better.
    @0.01, @0.005, and @0.001 denote the levels (standard deviations) of Gaussian noise.}
    \label{table:exp_noisy_recon_cd}
\end{center}
\end{table}


{\small
	\bibliographystyle{ieee_fullname}
	\bibliography{paper}
}